%% file: LDGnet.tex
\documentclass[10pt,twocolumn,journal]{IEEEtran}

\usepackage{epsfig}
\usepackage{epstopdf}
\usepackage{citesort}
\usepackage{amsmath}
\usepackage{amssymb}
\usepackage{color}
\usepackage{array}
\usepackage{multirow}
\usepackage{algorithm}
\usepackage{algorithmic}
\usepackage{rotating}
\usepackage{graphicx}
\newlength{\figurewidth}
\newlength{\smallfigurewidth}

\begin{document}
\title
{
Language-aware Domain Generalization Network for Cross-Scene Hyperspectral Image Classification
}

\author{%
Yuxiang Zhang,~\IEEEmembership{Student Member,~IEEE},
Mengmeng Zhang,
Wei Li,~\IEEEmembership{Senior Member,~IEEE},
Shuai Wang, \\
Ran Tao,~\IEEEmembership{Senior Member,~IEEE}
\thanks{%
	This work was supported by the National Key R\&D Program of China (2021YFB3900502), partly by National Natural Science Foundation of China under Grant 62001023. (corresponding author: M. Zhang; mengmengzhang@bit.edu.cn).  }
\thanks{%
Y. Zhang, M. Zhang, W. Li and R. Tao are with the School of Information and Electronics, Beijing Institute of Technology, and Beijing Key Laboratory of Fractional Signals and Systems, 100081 Beijing, China (e-mail: zyx829625@163.com, mengmengzhang@bit.edu.cn, liwei089@ieee.org, rantao@bit.edu.cn).
}
\thanks{
	S. Wang is with the Department of Chemistry, The University of Hong Kong, Hong Kong, China. (e-mail: shuaiw@connect.hku.hk)}
}

\maketitle
\thispagestyle{empty}
\pagestyle{empty}


\begin{abstract}
\input{abstract}
\end{abstract}

\begin{keywords}
Hyperspectral Image Classification,
Cross-Scene,
Domain Generalization,
Multiple-modality,
Natural Language Supervision,
Contrastive Learning.
\end{keywords}

\section{Introduction}
\input{introduction}

\section{Related Work}
\label{sec:related-works}

\input{related-works}

\section{Proposed Single-source Domain Expansion Network}
\label{sec:proposed}

\input{proposed}

\section{Experimental Results and Discussion}
\label{sec:results}
\input{results}

\section{Conclusions}
\label{sec:conclusions}
\input{conclusions}

\bibliographystyle{IEEEtran}
\bibliography{bibfile_zyx}

\end{document}

%% file: abstract.tex
Text information including extensive prior knowledge about land cover classes has been ignored in hyperspectral image classification (HSI) tasks. It is necessary to explore the effectiveness of linguistic mode in assisting HSI classification. In addition, the large-scale pre-training image-text foundation models have demonstrated great performance in a variety of downstream applications, including zero-shot transfer. However, most domain generalization methods have never addressed mining linguistic modal knowledge to improve the generalization performance of model. To compensate for the inadequacies listed above, a Language-aware Domain Generalization Network (LDGnet) is proposed to learn cross-domain invariant representation from cross-domain shared prior knowledge. The proposed method only trains on the source domain (SD) and then transfers the model to the target domain (TD). The dual-stream architecture including image encoder and text encoder is used to extract visual and linguistic features, in which coarse-grained and fine-grained text representations are designed to extract two levels of linguistic features. Furthermore, linguistic features are used as cross-domain shared semantic space, and visual-linguistic alignment is completed by supervised contrastive learning in semantic space. Extensive experiments on three datasets demonstrate the superiority of the proposed method when compared with state-of-the-art techniques.

%% file: introduction.tex
Inspired by the success of Deep Learning (DL) technology, remote sensing image classification based on advanced Convolutional Neural Networks (CNN) has received extensive attention, and achieved excellent performance, particularly in hyperspectral image (HSI) classification \cite{7428897,8467496,9775021}. However, most of CNN-based classification methods are only suitable for fixed scenes, that is, training samples and testing samples are independently and identically distributed. The acquisition process of HSI is inevitably affected by various factors, such as sensor nonlinearities, seasonal and weather conditions, which lead to variations in spectral reflectance between source domain (SD) and target domain (TD) of the same land cover classes. As a result, classification based on CNN has high generalization error and poor interpretation effect in the cross-scene classification task.

It is encouraging that transfer learning helps to solve this problem, where Domain Adaptation (DA) based approaches are widely used for cross-scene classification. Many DA methods have been developed for cross-scene classification from the perspective of DA, most notably related methods based on statistics, subspace learning, active learning, or deep learning. Among them, the maximum mean discrepancy (MMD) criterion \cite{6287330} was the earliest statistical technique used in the cross-scene interpretation. To address the issue of dynamic distribution adaptation in an adversarial network, Yu et al. developed the Dynamic Adversarial Adaptation Network (DAAN) \cite{yu2019transfer}. Class-wise distribution adaptation was created for HSI cross-scene classification \cite{liu2020class}, and the MMD method based on probability prediction was employed in an adversarial adaptation network to obtain more accurate feature alignment.

In recent years, a more difficult task setting, Domain Generalization (DG), has been developed in computer vision. The training samples of DA are labeled SD and unlabeled TD, that is, TD is accessed by the model during the training, while the training samples of DG are only labeled SD, and TD is not allowed to be accessed. The objective of DG is to learn a model from one or several different but related domains (i.e., diverse training datasets) that generalize well on TD \cite{wang2021generalizing}. Zhou et al. adversarially trained a transformation network for data augmentation instead of directly updating the inputs by gradient ascent \cite{zhou2020deep}. To replicate diverse photometric and geometric alterations in TD, Li et al. created the Progressive Domain Expansion Network (PDEN) \cite{li2021progressive}, a learning architecture that gradually produces multiple domains.

At present, most DG works consider how to learn domain invariant representation at the visual level from Multi-SDs or from a Single-SD. They have never tried to use language knowledge to assist visual representation learning and realize generalization by visual-linguistic alignment. In addition, it has been proved that language is helpful for visual representation learning in multi-modal learning \cite{radford2021learning,yu2022coca}. However, in HSI classification tasks, there is a lack of text information that is most able to reflect the prior knowledge of land cover classes. Therefore, how to construct a multi-modal learning framework suitable for HSI is worthy of in-depth study.

In order to address the aforementioned problems, a straightforward multi-modal DG framework for HSI, called Language-aware Domain Generalization Network (LDGnet), is proposed. Language knowledge is viewed as shared knowledge between SD and TD. The semantic space unifies visual modality and linguistic modality containing prior knowledge. A succinct visual-linguistic alignment strategy is designed to decrease the domain shift and increase the generalizability of the model. Specifically, LDGnet is split into three sections: image encoder, text encoder and visual-linguistic alignment. In the visual modality, the Deep residual 3D CNN network is employed by the image encoder to extract the visual features from the image patch. Second, to create semantic space in the linguistic modality, coarse- and fine-grained text descriptions are intended to supplement supervised signals with semantic information. The text encoder makes use of the universal language-model transformer. The core design of LDGnet is a visual-linguistic alignment strategy in which the semantic space is treated as a cross-domain shared space. In order to learn domain-invariant representations, supervised contrastive learning is utilized to project visual features into the semantic space.

The main contributions of this work are summarized as follows.

\begin{itemize}
	\item As far as we are aware, this is the first work to propose a language-guided framework for HSI classification, the introduction of a linguistic mode with prior knowledge as a supervised signal improved visual representation learning.
	
	\item The semantic space composed of linguistic features is regarded as cross-domain shared space, and the visual features are projected to the semantic space by visual-linguistic alignment to minimize the disparity across domains.
	
	\item In semantic space, coarse-grained and fine-grained text representations are designed, which enriches semantic supervised signals and promotes domain invariant representation learning.
\end{itemize}

The rest of the paper is organized as follows. Section II introduces relevant concepts of DG and Image-Text foundation models. Section III elaborates on the proposed LDGnet. The extensive experiments and analyses are presented in Section IV. Finally, conclusions are drawn in Section V.

%% file: related-works.tex
\begin{table}[]
	\begin{minipage}[!t]{\columnwidth}
	\renewcommand{\arraystretch}{1.3}
	\caption{\label{table:DADG}
	Comparison between domain adaptation and domain generalization}
			\begin{tabular}{cccc}
				\hline \hline
				Learning paradigm                                                              & Training data                              & Test data        & Test access       \\ \hline
				Domain adaptation                                                              & ${\cal S}^{src}, {\cal S}^{tar}$           & ${\cal S}^{tar}$ & $\surd$           \\ \hline
				\begin{tabular}[c]{@{}c@{}}Single-source \\ Domain generalization\end{tabular} & ${\cal S}^{src}$                             & ${\cal S}^{tar}$   & ${\rm{ \times }}$ \\ \hline
				\begin{tabular}[c]{@{}c@{}}Multi-source \\ Domain generalization\end{tabular}  & ${\cal S}^{1}, {\cal S}^{2}... {\cal S}^{n}$ & ${\cal S}^{n+1}$ & ${\rm{ \times }}$ \\ \hline \hline
			\end{tabular}
\end{minipage}
\end{table}

\begin{figure*}[tp] \small
	\vspace{-2em}
	\begin{center}
		\centering
		\epsfig{width=2.2\figurewidth,file=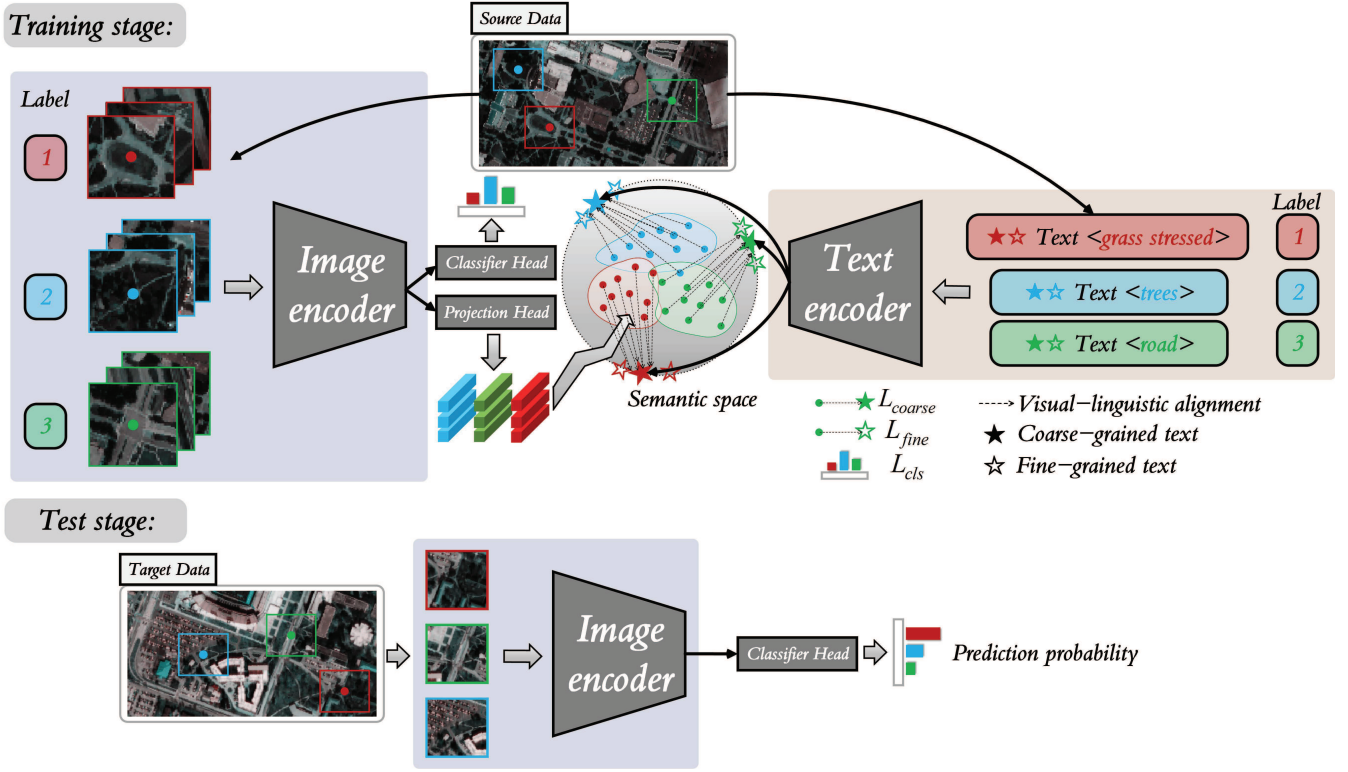}
		\caption{\label{fig:Flowchart}
			Flowchart of the proposed LDGnet. In training stage, the image encoder extracts visual features, and the text encoder extracts coarse-grained and fine-grained linguistic features to form a semantic space. Then, a visual-linguistic alignment is employed to reduce the gap between visual features and linguistic features by class, and finally the classification prediction probability of visual modality is output. In the test stage, the image encoder and classifier head is utilized for prediction of the image patch from TD. Red, green and blue represent three classes respectively. In the semantic space, dots represent visual features and pentagon represents linguistic features.}
	\end{center}
	\vspace{-2em}
\end{figure*}

\subsection{Domain Generalization (DG)}

DG is more challenging than DA, because DG aims to learn the model through SD data and does not need to access TD in the training phase. The model can be extended to TD in the inference stage. A comparison between DA and DG is listed in Table \ref{table:DADG}, where the ${\cal S}$ represents domain. The existing DG methods are divided into two categories: learning the domain invariant representation and data manipulation.

The key idea of the first category is to reduce the domain shift between multiple SD domain representations, which is mainly applied to multi-source DG. The most typical strategy is the explicit feature alignment. Some methods explicitly minimize the feature distribution divergence by minimizing MMD \cite{wang2020transfer}, second-order correlation \cite{sun2016deep}, Wasserstein distance \cite{zhou2020domain} of domains. Data manipulation is mainly applied to single-source DG. Such methods generally augment or generate out-of-domain samples related to SD, and then use these samples to train the model with the SD, and transfer to TD. Data generation creates diversified and abundant data to help generalization. For example, Variational Auto-encoder (VAE) \cite{kingma2013auto} and Generative Adversarial Networks (GAN) \cite{goodfellow2014generative} are often used for these purposes.

\subsection{Image-Text Foundation Models}

Recent researches have proposed image-text foundation models that learn strong joint representations between the two modalities through pretraining on large-scale image-text pairs and exhibit excellent transferability on some downstream visual and language tasks. The mainstream image-text foundation models are classified into two types based on their model architecture design: single-stream (early fusion of the two input modalities) and dual-stream (late fusion) \cite{gao2022pyramidclip}. The former one concatenates image and text input embeddings from VisualBERT \cite{li2019visualbert} and UNIMO \cite{li2021unimo} to model both image and text representations in a single transformer in a single semantic space. The latter, however, uses a decoupled image encoder and a text encoder for high-level representations, such as CLIP \cite{radford2021learning} and Coca \cite{yu2022coca}, to encode images and texts independently. Contrastive learning as training objectives is used by CLIP, DeCLIP, and Coca. The main principle of contrastive learning is to automatically build similar positive sample pairs and dissimilar negative sample pairs, with the goal of making positive sample pairs closer to one another in the projection space and negative sample pairs farther apart. In this paper, the dual-stream architecture and contrastive learning are used for LDGnet for flexible design and relatively cheap computation.

%% file: proposed.tex
\begin{figure*}[tp] \small
	\begin{center}
		\centering
		\epsfig{width=2\figurewidth,file=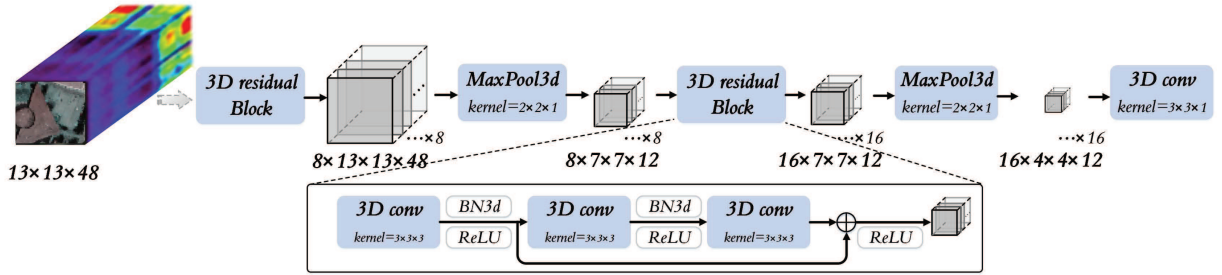}
		\caption{\label{fig:image}
			The flowchart of image encoder, where the 3D residual Block is the primary module.}
	\end{center}
	\vspace{-2em}
\end{figure*}

Assume that ${{\bf{X}}} = \left\{ {{\bf{x}}_i} \right\}_{i = 1}^{{N}}\in\mathbb{R}{^d}$ is the data from SD, and ${{\bf{Y}}}= \left\{ {{\bf{y}}_i} \right\}_{i = 1}^{{N}}$ is the corresponding class labels. Here, $d$ and $N$ denote the dimension of data and the number of source samples, respectively. The proposed LDGnet is broken into three parts: an image encoder, a text encoder, and a visual-linguistic alignment. The flowchart for each part is illustrated in Fig.~\ref{fig:Flowchart}. The sample of 13$\times $13$\times d$ spatial patch in HSI is selected from SD, and each one is assigned a coarse-grained text and two fine-grained texts based on the class of the patch. In the visual mode, the image is delivered to the image encoder to extract features, and then the classification head $Cls$ is used to calculate the cross entropy loss, and the projection head $Proj$ outputs visual features for visual-linguistic alignment. To extract the linguistic features and create the cross-domain shared semantic space, the text is passed to the text encoder. Finally, supervised contrastive learning is employed for visual-linguistic alignment to align image and text by class.

\subsection{Image encoder}

Given that HSI is a 3D data cube made up of hundreds of spectral channels with a wealth of spatial and spectral information, the Deep residual 3D CNN network is used to extract spatial-spectral features, as shown in Fig.~\ref{fig:image}. The image encoder is made up of two 3D residual Block-MaxPool3d modules and one Conv3d module, where the 3D residual Block is the core module for extracting robust visual features. The 3D residual Block is composed of two Conv3d-BN3d-ReLUs and one Conv3d in series. The output of the first Conv3d-BN3d-ReLU and the output of Conv3d are used for residual connection. The extracted spatial-spectral features are sent to the classification head and the cross entropy is calculated with the ground truth. At the same time, it is fed into the projection head to obtain the visual features for feature alignment of visual modes and linguistic modes. The cross-entropy loss is defined as,
\begin{equation}\label{eq:11}
{{\cal L}_{ce}}\left( {{{\bf{p}}_i},{{\bf{y}}_i}} \right) =  - \sum\limits_c {y_i^c\log p_i^c} 
\end{equation}
where ${\bf{y}}_i$ is the one-hot encoding of the label information of ${{\bf{x}}_i}$, $c$ is the index of class, and ${{\bf{p}}_i}$ is the predicted probability output obtained by $Cls$. Therefore, the classification loss for SD is defined as,
\begin{equation}\label{eq:12}
{{{\cal L}_{SD}}\left( {{\bf{X}},{\bf{Y}}} \right) = \frac{1}{N}\sum\limits_i {{{\cal L}_{ce}}\left( {C\left( {{{\bf{x}}_i}} \right),{{\bf{y}}_i}} \right).} }
\end{equation}

\subsection{Text encoder}

The text encoder is a language-model transformer \cite{vaswani2017attention} modified in accordance with Radford et al. \cite{radford2019language}. The text encoder start with a base size of a 33M-parameter, 3-layer, 512-wide model with 8 attention heads. Similar to CLIP, the text is represented by the transformer using a lower-cased byte pair encoding (BPE) with a vocabulary size of 49,152 \cite{sennrich2016neural}. The maximum sequence length is limited to 76 for computational efficiency. 
This linguistic feature is then normalized by layer and linearly projected into the semantic space.

\begin{figure}[tp] \small
	\begin{center}
		\centering
		\epsfig{width=1\figurewidth,file=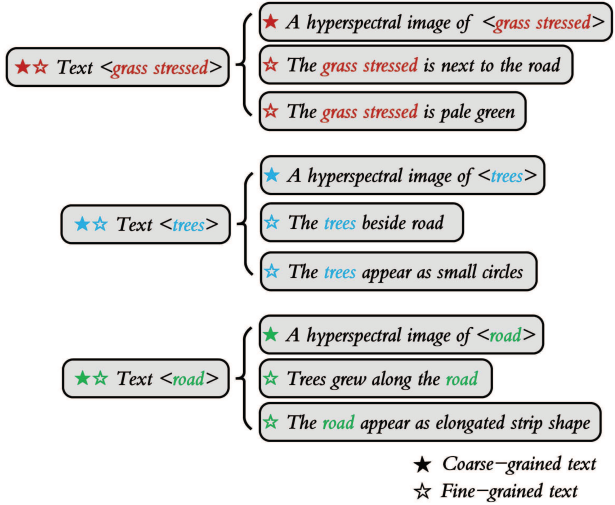}
		\caption{\label{fig:prompt}
			Examples of coarse-grained and fine-grained text for three classes.}
	\end{center}
	\vspace{-2em}
\end{figure}

According to the prior knowledge of land cover classes in SD scene, coarse-grained and fine-grained text descriptions are assigned to each land cover classes. As shown in Fig.~\ref{fig:prompt}, \textcolor[rgb]{0.3,0.3,0.3}{\sffamily A hyperspectral image of \textless class name\textgreater} is used as a template to construct coarse-grained text descriptions for each class in a cloze way, the solid pentagon shown in Fig.~\ref{fig:prompt}. As for fine-grained text, combined with prior knowledge, the color, shape, distribution and adjacency relationship are described manually, such as \textcolor[rgb]{0.3,0.3,0.3}{\sffamily The grass stress is pale green}, \textcolor[rgb]{0.3,0.3,0.3}{\sffamily The trees beside road} and \textcolor[rgb]{0.3,0.3,0.3}{\sffamily The road appears as elongated strip shape}. Each land cover class sets two fine-grained text, the hollow pentagon shown in Fig.~\ref{fig:prompt}.

\begin{figure*}[htp] \small
	\setlength{\tabcolsep}{0.1em}
	\begin{center}
		\begin{tabular}{cccc}
			\epsfig{width=0.55\figurewidth,file=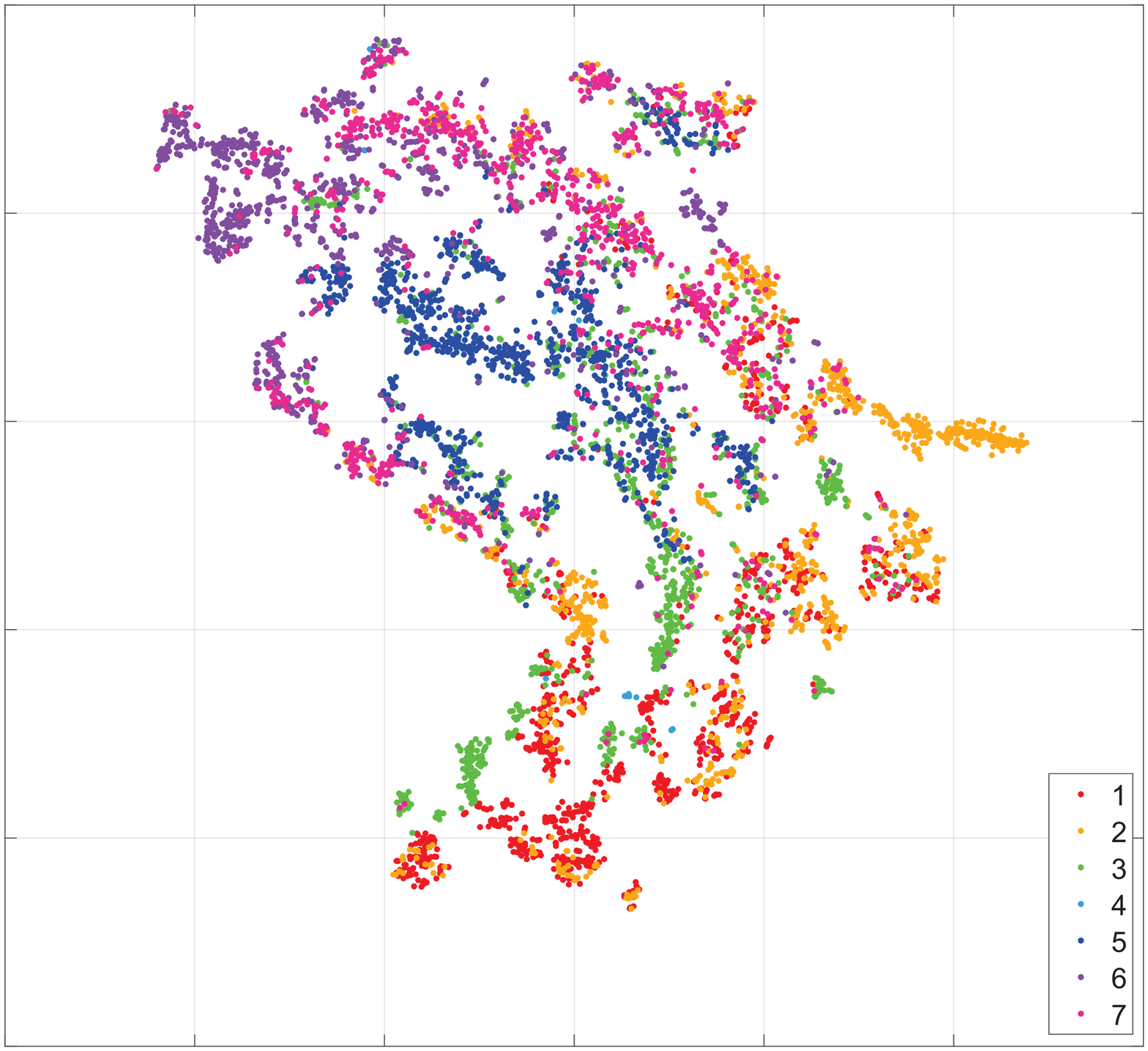} &
			\epsfig{width=0.55\figurewidth,file=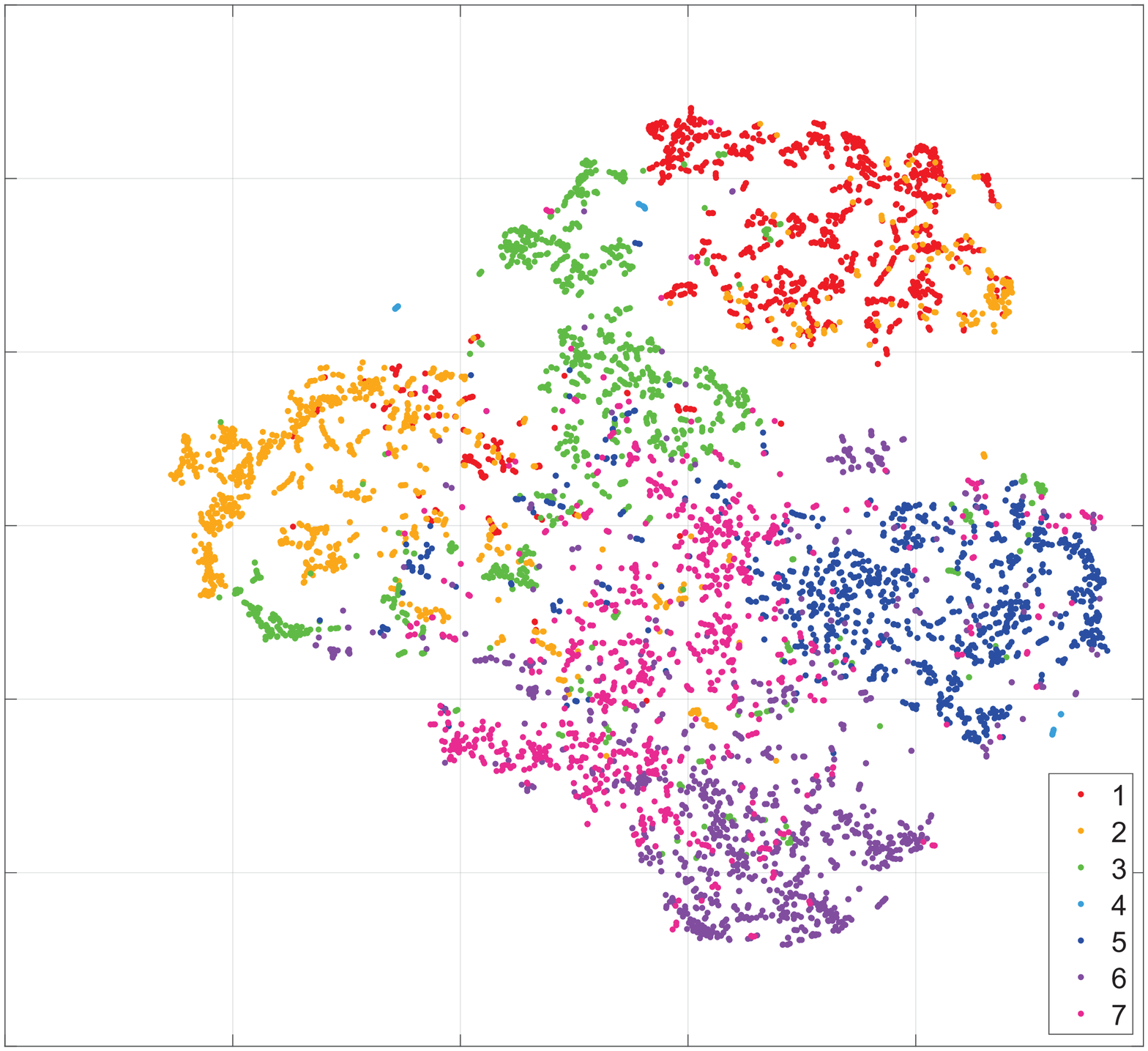}&
			\epsfig{width=0.55\figurewidth,file=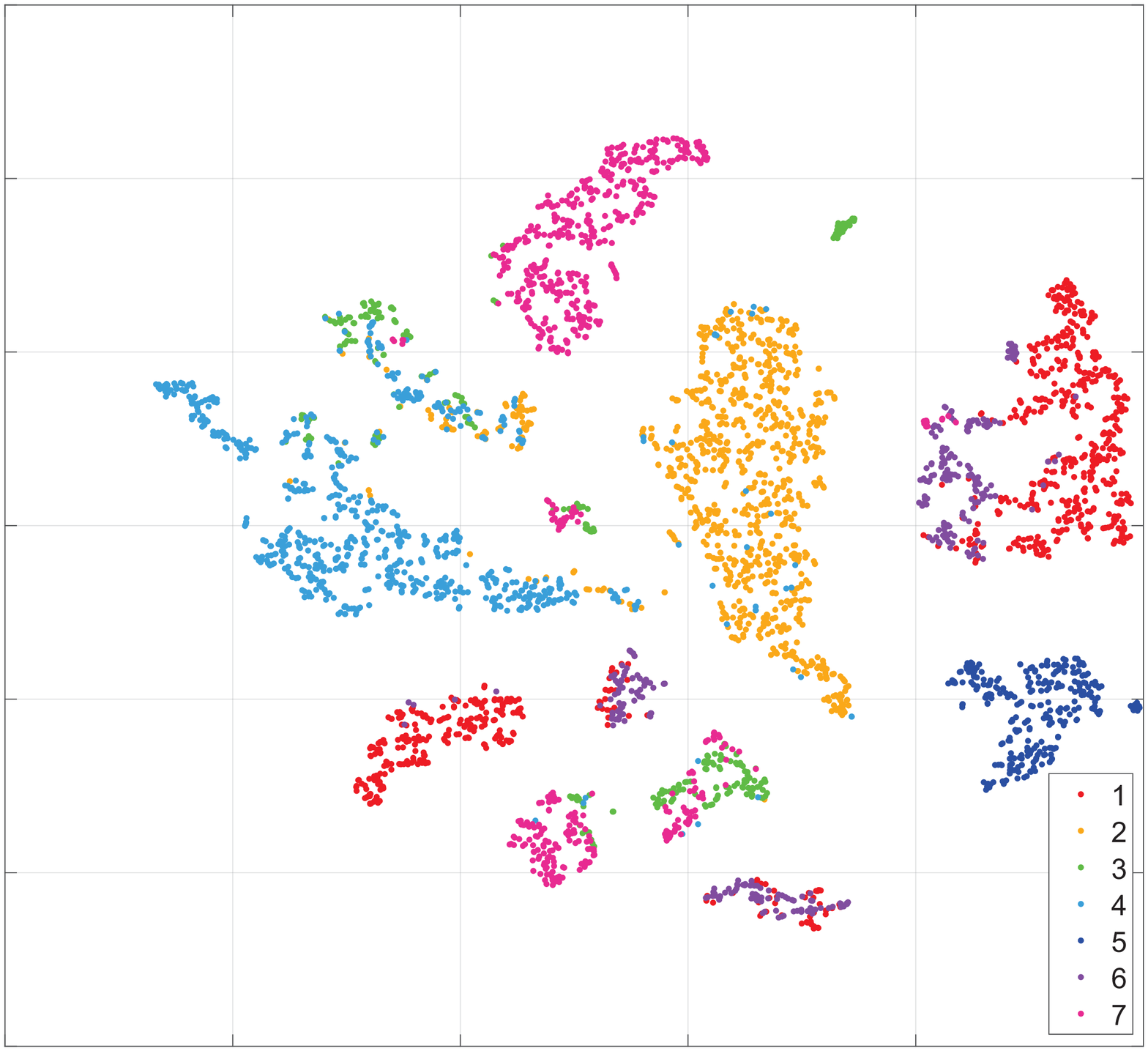}&
			\epsfig{width=0.55\figurewidth,file=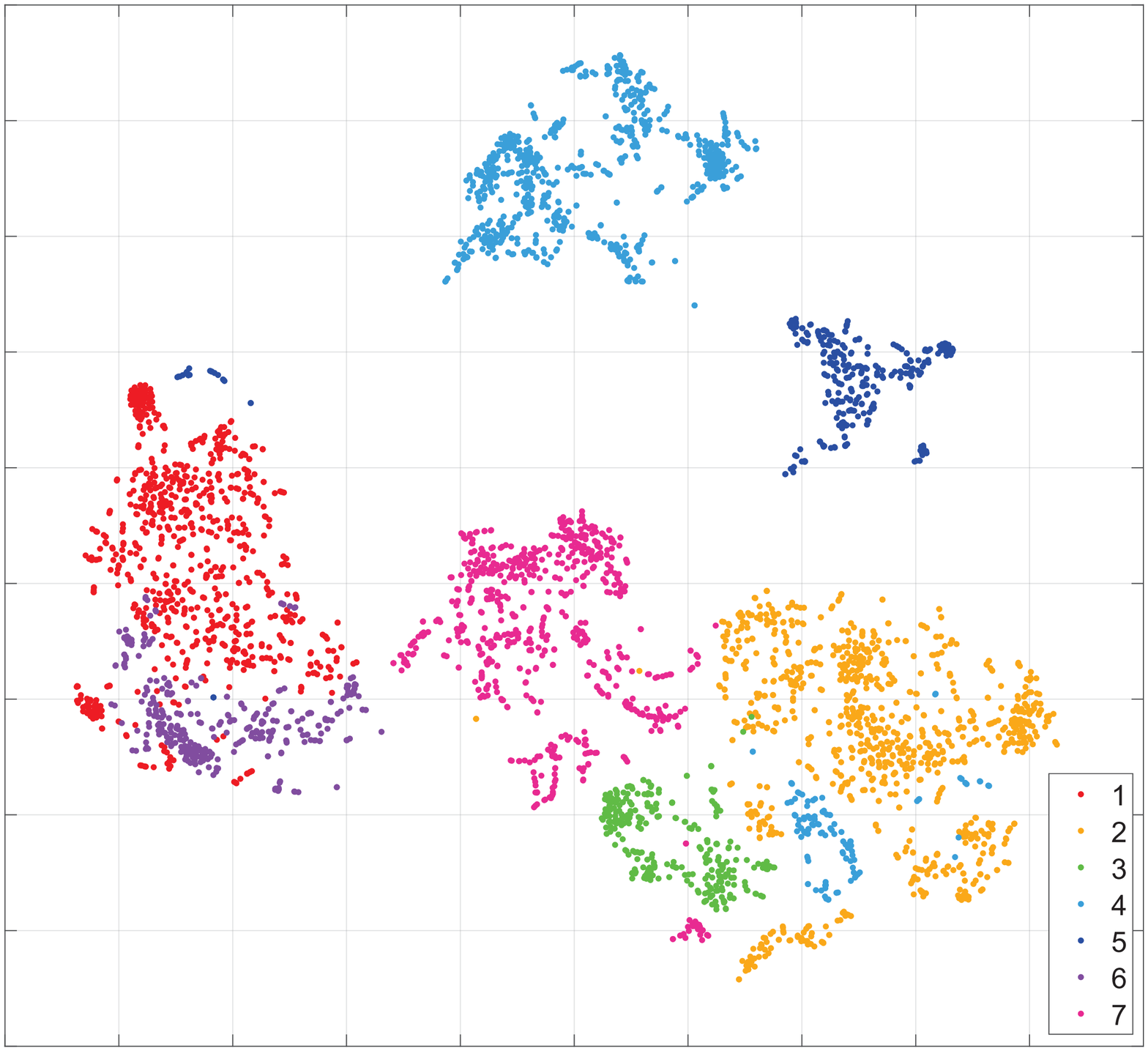} \\
			(a)  Original samples from TD & (b) TD features by LDGnet & (c) Original samples from TD  & (d) TD features by LDGnet\\  [0.5em]
		\end{tabular}
		\caption{\label{fig:analysis}
			Class separability of the proposed LDGnet using three datasets, where (a) and (b) are the Houston 2018 data, and (c) and (d) are the Pavia Center data. Best looking in a$\to$b and c$\to$d.}
	\end{center}
\end{figure*}

\subsection{Visual-linguistic alignment}

The linguistic features obtained by text encoder construct semantic space as a cross-domain shared space. To achieve the alignment of visual features and linguistic features by class, supervised contrastive learning is performed, and the domain invariant representation driven by language is developed to generalize the model to TD. Firstly, a supervised contrastive learning is defined as,
\begin{equation}\label{eq:1}
{{\cal L}_{supcon}} =  - \sum\limits_{i = 0}^N {\frac{1}{{|P(i)|}}} \sum\limits_{p \in P(i)} {\log } \frac{{\exp \left( {{{{\bf{x}}_i^T{\bf{x}}_p^ + } \mathord{\left/
					{\vphantom {{{\bf{x}}_i^T{\bf{x}}_p^ + } \tau }} \right.
					\kern-\nulldelimiterspace} \tau }} \right)}}{{\sum\limits_{a \in A(i)} {\exp \left( {{{{\bf{x}}_i^T{\bf{x}}_a^ - } \mathord{\left/
						{\vphantom {{{\bf{x}}_i^T{\bf{x}}_a^ - } \tau }} \right.
						\kern-\nulldelimiterspace} \tau }} \right)} }}
\end{equation}
where for each embedding feature ${{\bf{x}}_i}$ in minibatch, ${P(i)}$ and ${A(i)}$ are the positive and negative sample sets, ${|P(i)|}$ is the number of positive samples, ${{\bf{x}}_p^ + }$ and ${{\bf{x}}_a^ - }$ are one of the positive and negative samples. 

Given all image-text pairs in a minibatch, where image and text are one-to-one and have the same class label. The alignment losses ${{\cal L}_{supcon}}$ of image to text and text to image are calculated,
\begin{equation}\label{eq:2}
\begin{array}{l}
{{\cal L}_{coarse}} =  - \sum\limits_{i = 0}^N {\frac{1}{{|P(i)|}}} (\sum\limits_{p \in {P_l}(i)} {\log } \frac{{\exp ({{{\bf{v}}_i^T{\bf{l}}_p^ + } \mathord{\left/
				{\vphantom {{{\bf{v}}_i^T{\bf{l}}_p^ + } \tau }} \right.
				\kern-\nulldelimiterspace} \tau })}}{{\sum\limits_{a \in {A_l}(i)} {\exp ({{{\bf{v}}_i^T{\bf{l}}_a^ - } \mathord{\left/
					{\vphantom {{{\bf{v}}_i^T{\bf{l}}_a^ - } \tau }} \right.
					\kern-\nulldelimiterspace} \tau })} }}\\
\begin{array}{*{20}{c}}
{}&{}&{}
\end{array} + \sum\limits_{p \in {P_v}(i)} {\log } \frac{{\exp {{({\bf{l}}_i^T{\bf{v}}_p^ + } \mathord{\left/
				{\vphantom {{({\bf{l}}_i^T{\bf{v}}_p^ + } \tau }} \right.
				\kern-\nulldelimiterspace} \tau })}}{{\sum\limits_{a \in {A_v}(i)} {\exp ({{{\bf{l}}_i^T{\bf{v}}_a^ - } \mathord{\left/
					{\vphantom {{{\bf{l}}_i^T{\bf{v}}_a^ - } \tau }} \right.
					\kern-\nulldelimiterspace} \tau })} }})
\end{array}
\end{equation}
where for each visual feature ${{\bf{v}}_i}$ and linguistic feature ${{\bf{l}}_i}$ in minibatch, ${{P_v}(i)}$ and ${{A_v}(i)}$ are the positive and negative sample sets of visual feature, ${{P_l}(i)}$ and ${{A_l}(i)}$ are the positive and negative sample sets of linguistic feature, ${|{P_v}(i)|=|{P_l}(i)|=|P(i)|}$. In addition, the temperature parameter $\tau$, which governs the range of the logits in the softmax, is explicitly optimized as a log-parameterized multiplicative scalar during training. ${\cal L}_{coarse}$ represents the alignment loss of image to coarse-grained text. The alignment loss of image to fine-grained text ${\cal L}_{fine}$ is similar to ${\cal L}_{coarse}$, with the difference being that linguistic features require two fine-grained linguistic features. In the process of optimizing, the visual features and linguistic features belonging to the same class are put into ${{P_v}(i)}$ and ${{P_l}(i)}$, and the features outside the class are put into ${{A_v}(i)}$ and ${{A_l}(i)}$. The image encoder and text encoder are optimized by ${\cal L}_{coarse}$ and ${\cal L}_{fine}$ to bring features from the same class closer together and samples from different classes farther apart. This allows image encoder to learn class-specific domain invariant representations.

In LDGnet, the image encoder and text encoder are jointly trained. Integrating above loss functions, the total loss of LDGnet is defined as,
\begin{equation}\label{eq:17}
{{\cal L}_{total}} = {{\cal L}_{SD}} + \lambda \left( {\left( {1 - \alpha } \right){{\cal L}_{coarse}} + \alpha {{\cal L}_{fine}}} \right)
\end{equation}
where $\lambda$ is a hyper-parameter for balancing the alignment loss, and $\alpha$ is employed to control the contribution of both coarse- and fine-grained linguistic features. In the test phase, only image encoder and classification header are used to predict TD.

\subsection{Generalization Performance of LDGnet}

Language is employed in LDGnet as an additional supervised signal. Visual and linguistic features are aligned in a cross-domain shared semantic space, and class-wise domain invariant representations are learned from multi-modal features. Two TDs (Houston 2018 data and Pavia Center data) are inferred using fully trained LDGnet, and the visualization of class separability in the original space and semantic space are shown in Fig. \ref{fig:analysis}. The inter-class distribution is messed up in Fig. \ref{fig:analysis}(a) and (c), and after features are embedded in the semantic space, the separability is greatly increased. For instance, in the 1-st class of (b) and 3-rd class of (d), the aggregate of features belonging to the same class is greatly improved.

%% file: results.tex
Experiments using three cross-scene HSI datasets, i.e., the Houston dataset, Pavia dataset, and GID (Gaofen Image Dataset) dataset, are conducted to validate the proposed LDGnet. For comparison algorithms, several state-of-the-art transfer learning algorithms are used, including DA methods, Dynamic Adversarial Adaptation Network (DAAN) \cite{yu2019transfer}, Deep Subdomain Adaption Network (DSAN) \cite{2020Deepsub}, Multi-Representation Adaptation Network (MRAN) \cite{ZHU2019214} and Heterogeneous Transfer CNN (HTCNN) \cite{2019Heterogeneous}, DG methods, Progressive Domain Expansion Network (PDEN) \cite{li2021progressive}, LDSDG (Learning to Diversify for Single Domain Generalization) \cite{wang2021learning} and Style-Agnostic Network (SagNet) \cite{nam2021reducing}. The class-specific accuracy (CA), the overall accuracy (OA) and the Kappa coefficient (KC) are employed to evaluate the classification performance.

\subsection{Experimental Data}
\begin{figure}[tp]
	\begin{center}
		\begin{tabular}{cc}
			\epsfig{width=0.5\figurewidth,file=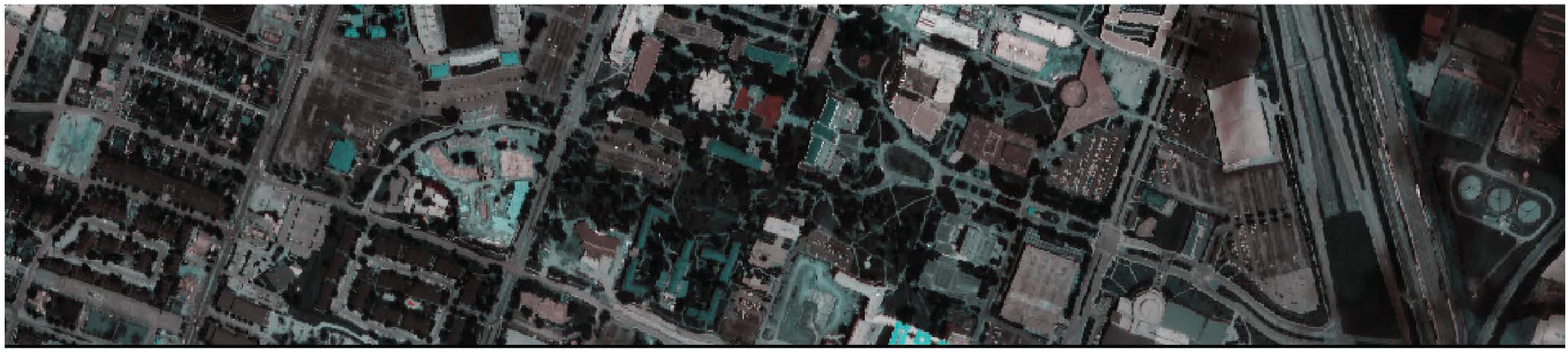} &
			\epsfig{width=0.5\figurewidth,file=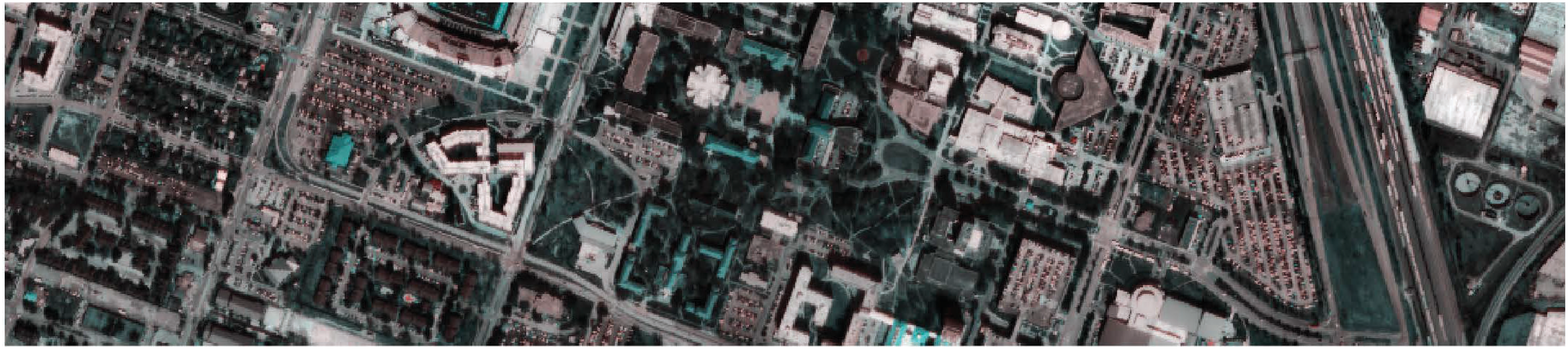} \\
			(a) & (b) \\
			\epsfig{width=0.5\figurewidth,file=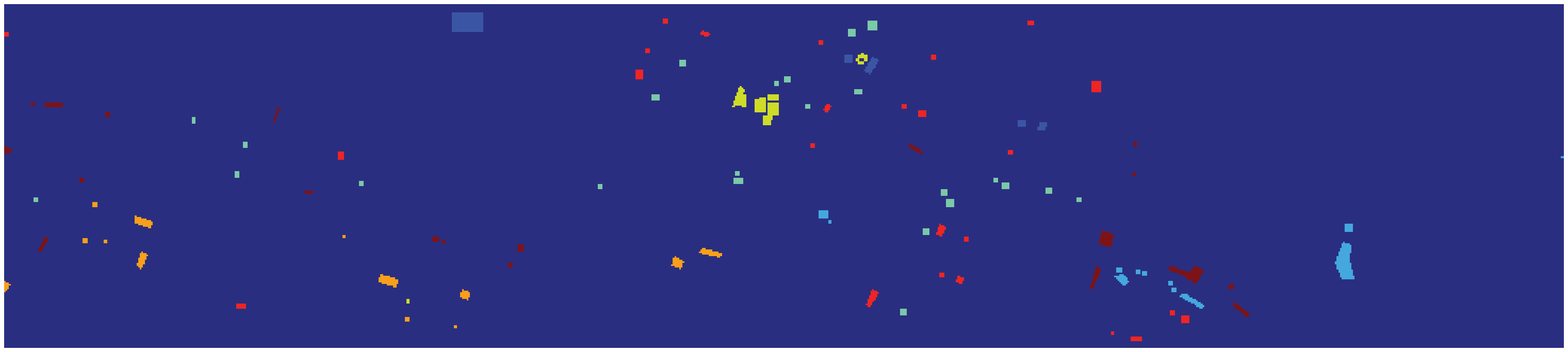} &
			\epsfig{width=0.5\figurewidth,file=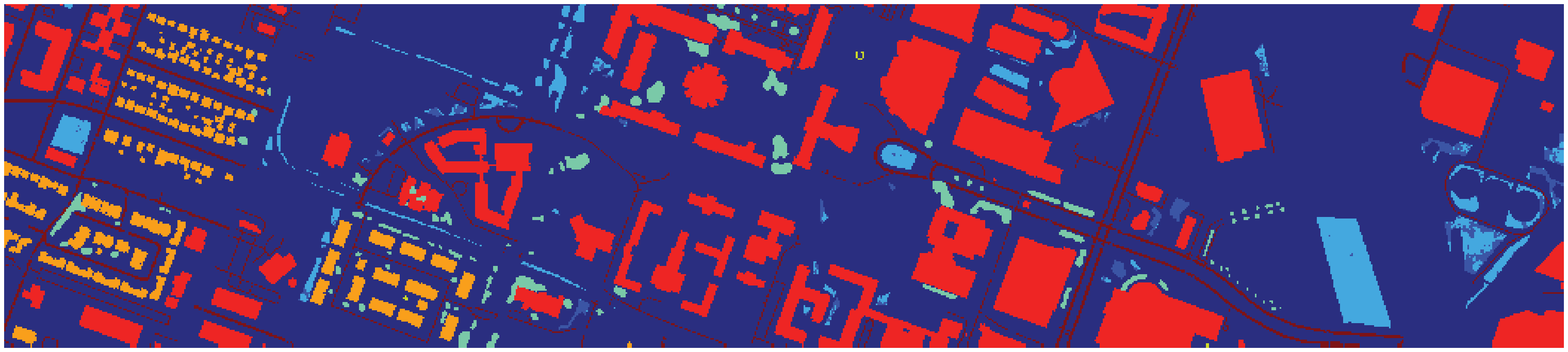} \\
			(c) & (d)   \\
		\end{tabular}
		\begin{tabular}{cc}
			\epsfig{width=\figurewidth,file=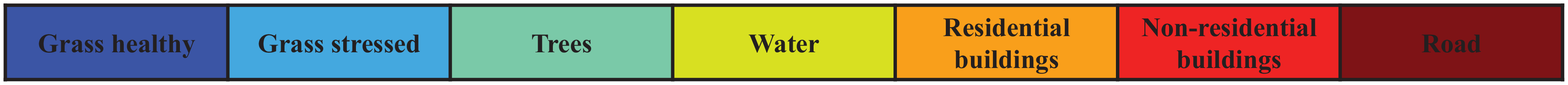}\\
		\end{tabular}
	\end{center}
	\caption{\label{fig:Houston_fg}
		Pseudo-color image and ground truth map of Houston dataset: (a) Pseudo-color image of Houston 2013, (b) Pseudo-color image of Houston 2018, (c) Ground truth map of Houston 2013, (d) Ground truth map of Houston 2018. }
\end{figure}

\textbf{Houston dataset}: The dataset includes Houston 2013 \cite{2014Hyperspectral} and Houston 2018 \cite{20182018} scenes, which were obtained by different sensors on the University of Houston campus and its vicinity in different years. The Houston 2013 dataset is composed of 349$\times$1905 pixels, including 144 spectral bands, the wavelength range is 380-1050nm, and the image spatial resolution is 2.5m. The Houston 2018 dataset has the same wavelength range but contains 48 spectral bands, and the image has a spatial resolution of 1m. There are seven consistent classes in their scene. We extract 48 spectral bands (wavelength range 0.38$\sim$1.05um) from Houston 2013 scene corresponding to Houston 2018 scene, and select the overlapping area of 209$\times$955. The classes and the number of samples are listed in Table \ref{table:Houston_samples}. Additionally, their false-color and ground truth maps are shown in Fig. \ref{fig:Houston_fg}.

\textbf{Pavia dataset}: The Pavia dataset include University of Pavia (UP) and Pavia Center (PC). The PC has 1096$\times$715 pixels and 102 bands. The UP has 103 spectral bands, 610$\times$340 pixels and 1.3 m spatial resolution, where the last band was removed to ensure the same number of spectral bands as PC. They all have the same seven classes and the name of land cover classes and the number of samples are listed in Table \ref{table:pavia_samples}.

\begin{figure}[tp]
	\begin{center}
		\begin{tabular}{cccc}
			\begin{sideways}\epsfig{width=0.35\figurewidth,file=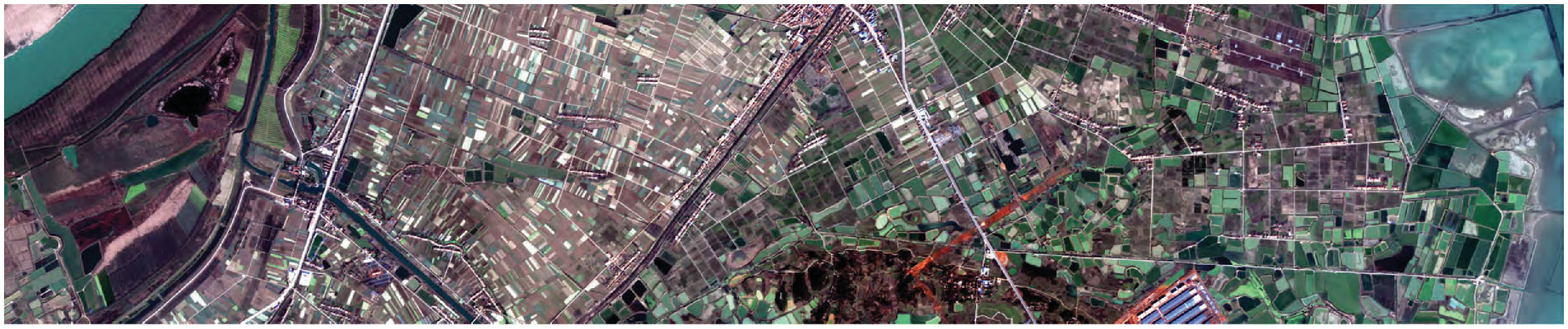}\end{sideways} &			\begin{sideways}\epsfig{width=0.35\figurewidth,file=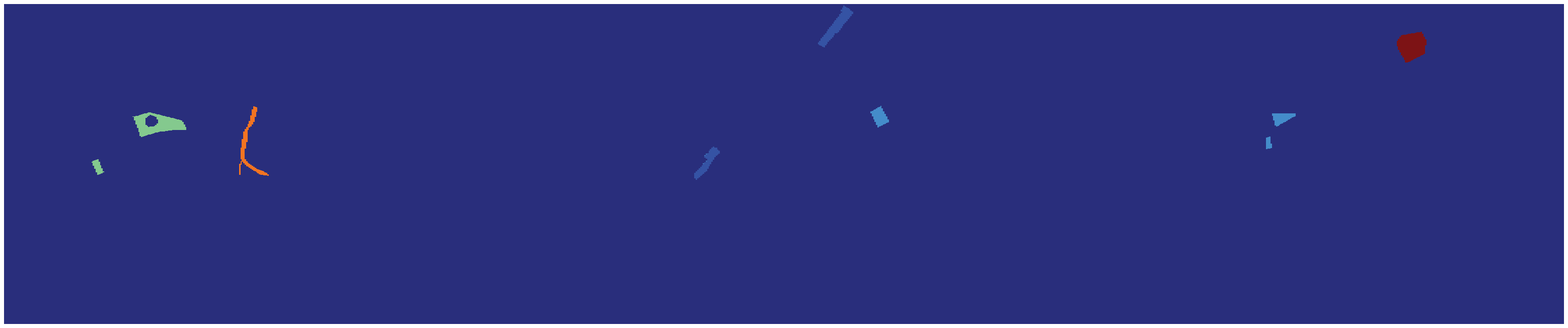} \end{sideways}&
			\epsfig{width=0.3\figurewidth,file=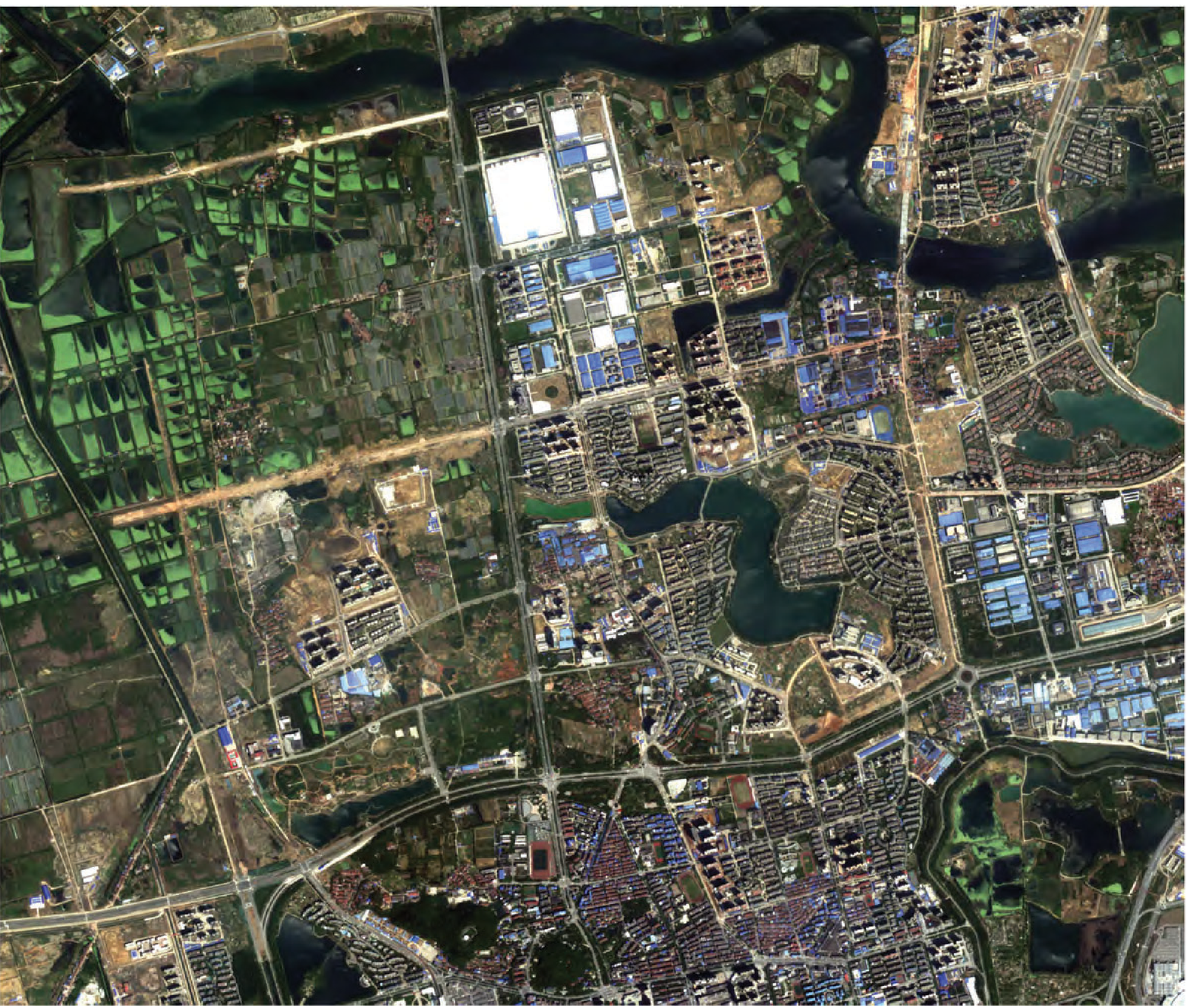} &
			\epsfig{width=0.3\figurewidth,file=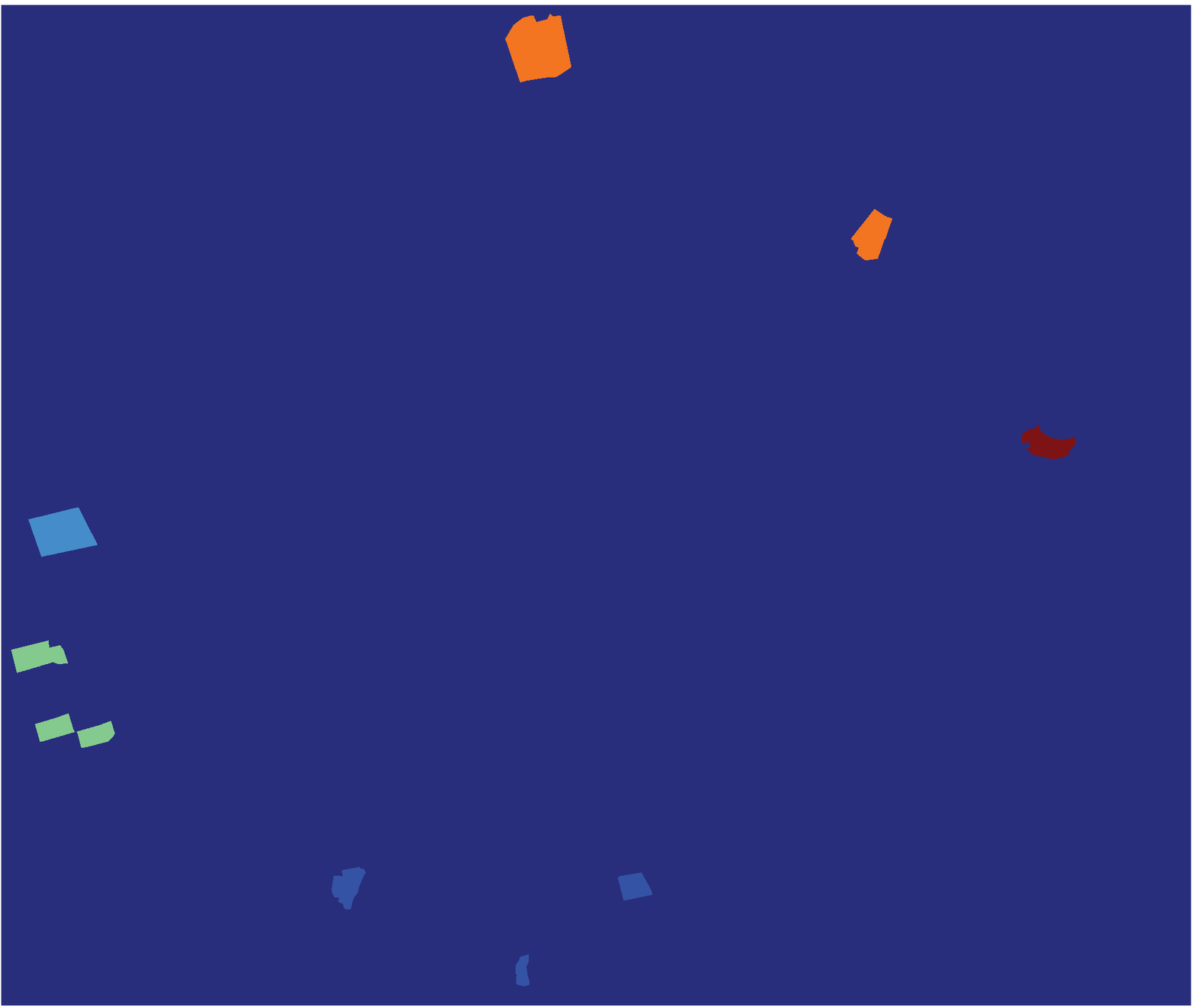} \\
				(a) & (b) & (c) & (d)  
		\end{tabular}
		\begin{tabular}{cc}
			\epsfig{width=1\figurewidth,file=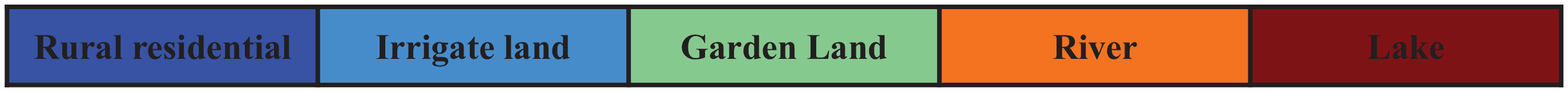}\\
		\end{tabular}
	\end{center}
	\caption{\label{fig:GID_fg}
		Pseudo-color image and ground truth map of GID dataset: (a) Pseudo-color image of GID-nc, (b) Ground truth map of GID-nc, (c) Pseudo-color image of GID-wh, (d) Ground truth map of GID-wh.}
\end{figure}

\textbf{GID dataset}: GID dataset is constructed by Wuhan University \cite{tong2020land}, which contains multispectral images (MSI) taken at different times in many regions of China. We selected GID-nc shot in Nanchang, Jiangxi Province, on January 3, 2015 as the source domain, and GID-wh shot in Wuhan, Hubei Province, on April 11, 2016 as the target domain. GID-nc consists of 900$\times$4400 pixels, including blue (0.45-0.52um), green (0.52-0.59um), red (0.63-0.69um) and near infrared (0.77-0.89um) bands, and the spatial resolution is 4m. GID-wh also has the same spatial and spectral resolution, but it is composed of 1600$\times$1900 pixels. They have the same five classes, as listed in Table \ref{table:GID}. The false-color images and ground-truth maps are shown in Fig. \ref{fig:GID_fg}.

\begin{table}[tp]
	\caption{\label{table:Houston_samples}
		Number of source and target samples for the Houston dataset.}
	{
		\begin{center}
			\begin{tabular}{|c|c|c|c|}
				\hline
				\hline
				\multicolumn{2}{|c|}{Class} &\multicolumn{2}{c|}{Number of Samples} \\
				\hline
				\multirow{2}{*}{No.} &\multirow{2}{*}{Name}  & Houston 2013 & Houston 2018\\
				{~}  & {~}  & {(Source)} &  {(Target)} \\
				\hline
				1     & Grass healthy & 345   & 1353 \\
				2     & Grass stressed & 365   & 4888 \\
				3     & Trees & 365   & 2766 \\
				4     & Water & 285   & 22 \\
				5     & Residential buildings & 319   & 5347 \\
				6     & Non-residential buildings & 408   & 32459 \\
				7     & Road  & 443   & 6365 \\
				\hline
				\multicolumn{2}{|c|}{Total} & 2530& 53200\\
				\hline \hline
			\end{tabular}
	\end{center}}
\vspace{-2em}
\end{table}

\begin{table}[tp]
	\caption{\label{table:pavia_samples}
		Number of source and target samples for the Pavia dataset.}
	{
		\begin{center}
			\begin{tabular}{|c|c|c|c|}
				\hline
				\hline
				\multicolumn{2}{|c|}{Class} &\multicolumn{2}{c|}{Number of Samples} \\
				\hline
				\multirow{2}{*}{No.} &\multirow{2}{*}{Name}  & UP & PC\\
				{~}  & {~}  & {(Source)} &  {(Target)} \\
				\hline
				1 & Tree      & 3064  & 7598 \\
				2 & Asphalt   & 6631  & 9248 \\
				3 & Brick     & 3682  & 2685 \\
				4 & Bitumen   & 1330  & 7287 \\
				5 & Shadow    & 947   & 2863 \\
				6 & Meadow    & 18649 & 3090 \\
				7 & Bare soil & 5029  & 6584 \\
				\hline
				\multicolumn{2}{|c|}{Total} & 39332& 39355\\
				\hline \hline
			\end{tabular}
	\end{center}}
\end{table}

\begin{table}[tp]
	\caption{\label{table:GID}
		Number of source and target samples for the GID dataset.}
	{
		\begin{center}
			\begin{tabular}{|c|c|c|c|}
				\hline
				\hline
				\multicolumn{2}{|c|}{Class} &\multicolumn{2}{c|}{Number of Samples} \\
				\hline
				\multirow{2}{*}{No.} &\multirow{2}{*}{Name}  & GID-nc & GID-wh\\
				{~}  & {~}  & {(Source)} &  {(Target)} \\
				\hline
				1     & Rural residential & 5495  & 4729 \\
				2     & Irrigate land & 3643   & 5643 \\
				3     & Garden Land & 6171   & 6216 \\
				4     & River& 2858   & 11558 \\
				5     & Lake & 5172  & 2666 \\
				\hline
				\multicolumn{2}{|c|}{Total} & 23339 & 30812\\
				\hline \hline
			\end{tabular}
	\end{center}}
\end{table}

\subsection{Experimental Setting}
LDGnet is implemented on the Pytorch platform. The input is configured with a patch size of 13$\times$13. The image encoder and text encoder are optimized via Adaptive Moment Estimation (Adam). The default value for ${\ell _2}$-norm regularization of all modules is set to 1e-4 for weight decay. As the initialization of the text encoder, load the \emph{ViT-B-32.pt} that was completed by CLIP pretraining. \textcolor[rgb]{0.3,0.3,0.3}{\sffamily A hyperspectral image of \textless class name\textgreater} is used as a template to generate coarse-grained text descriptions for each class. Furthermore, we artificially set the fine-grained text descriptions of three datasets, as seen Tables \ref{table:Houston_text}-\ref{table:GID_text}.


\begin{table}[]
		\caption{\label{table:Houston_text}
		Fine-grained text descriptions for the Houston dataset.}
	\begin{tabular}{|c|l|}
		\hline \hline
		Class name                                                                           & \multicolumn{1}{c|}{Fine-grained text}                                                                              \\ \hline
		\multirow{2}{*}{Grass healthy}                                                                         & The grass healthy is next to the road                                                                               \\ 
		& The grass healthy is dark green                                                                                     \\ \hline
		\multirow{2}{*}{Grass stressed}                                                      & The grass stressed is next to the road                                                                              \\ \cline{2-2} 
		& The grass stressed is pale green                                                                                    \\ \hline
		\multirow{2}{*}{Trees}                                                               & \begin{tabular}[c]{@{}l@{}}The trees are next to residential buildings \\ or non-residential buildings\end{tabular} \\ \cline{2-2} 
		& The trees appear as small circles                                                                                   \\ \hline
		\multirow{2}{*}{Water}                                                               & The water has a smooth surface                                                                                      \\ \cline{2-2} 
		& The water appears dark blue or black                                                                                \\ \hline
		\multirow{2}{*}{\begin{tabular}[c]{@{}c@{}}Residential\\ buildings\end{tabular}}     & Residential buildings are densely packed                                                                            \\ \cline{2-2} 
		& Residential buildings appear as small blocks                                                                        \\ \hline
		\multirow{2}{*}{\begin{tabular}[c]{@{}c@{}}Non-residential\\ buildings\end{tabular}} & \begin{tabular}[c]{@{}l@{}}The shapes of the non-residential buildings are \\ inconsistent\end{tabular}                                                          \\ \cline{2-2} 
		& Non-residential buildings appear as large blocks                                                                    \\ \hline
		\multirow{2}{*}{Road}                                                                & Trees grew along the road                                                                                           \\ \cline{2-2} 
		& The road appear as elongated strip shape                                                                            \\ \hline \hline
	\end{tabular}
\end{table}

%
%
%
%
%
%

\begin{table}[]
	\caption{\label{table:Pavia_text}
		Fine-grained text descriptions for the Pavia dataset.}
	\begin{center}
	\begin{tabular}{|c|l|}
		\hline \hline
		Class name                 & \multicolumn{1}{c|}{Fine-grained text}           \\ \hline
		\multirow{2}{*}{Trees}     & The trees beside road                            \\ \cline{2-2} 
		& The trees appear as small circles                \\ \hline
		\multirow{2}{*}{Asphalt}   & Trees grew along the asphalt road                \\ \cline{2-2} 
		& The asphalt road appear as elongated strip shape \\ \hline
		\multirow{2}{*}{Brick}     & Brick is a kind of road material                 \\ \cline{2-2} 
		& Bricks are generally arranged in strips          \\ \hline
		\multirow{2}{*}{Bitumen}   & Bitumen is a material for building surfaces      \\ \cline{2-2} 
		& Bitumen is a widely used waterproof material     \\ \hline
		\multirow{2}{*}{Shadow}    & The shadows next to buildings                    \\ \cline{2-2} 
		& The shadow appears black                         \\ \hline
		\multirow{2}{*}{Meadow}    & The surface of the meadow is green with grass    \\ \cline{2-2} 
		& The meadow is pale green                         \\ \hline
		\multirow{2}{*}{Bare soil} & No grass on the surface of bare soil             \\ \cline{2-2} 
		& The bare soil appears grayish-black color        \\ \hline \hline
	\end{tabular}
	\end{center}
\end{table}

%
%
%
%
%
%


\begin{table}[]
	\caption{\label{table:GID_text}
		Fine-grained text descriptions for the GID dataset.}
	\begin{center}
	\begin{tabular}{|c|l|}
		\hline \hline
		Class name                         & \multicolumn{1}{c|}{Fine-grained text}                                                                              \\ \hline
		\multirow{2}{*}{Rural residential} & Rural residential buildings are densely packed                                                                      \\ \cline{2-2} 
		& Rural residential buildings appear as small blocks                                                                  \\ \hline
		\multirow{2}{*}{Irrigated land}    & \begin{tabular}[c]{@{}l@{}}Irrigated land is land that depends on irrigation\\ for normal productivity\end{tabular} \\ \cline{2-2} 
		& Irrigated land with crops and plenty of water                                                                       \\ \hline
		\multirow{2}{*}{Garden Land}       & The soil is exposed in garden land                                                                                  \\ \cline{2-2} 
		& Vegetation is sparse on garden land                                                                                 \\ \hline
		\multirow{2}{*}{River}             & The river runs through the land very long                                                                           \\ \cline{2-2} 
		& The river appears as long strip                                                                                     \\ \hline
		\multirow{2}{*}{Lake}              & The lake is static and does not flow                                                                                \\ \cline{2-2} 
		& The lake has no fixed or regular shape                                                                              \\ \hline \hline
	\end{tabular}
	\end{center}
\end{table}
%
%
%
%

\begin{table}[]
	\caption{\label{tab:lr}
		Overall classification accuracy (\%) with different base learning rates ${\eta}$ for the proposed LDGnet using the three experimental data.}
	\centering
	\begin{tabular}{|c|c|c|c|c|c|}
		\hline\hline
		\multirow{2}{*}{Target scene} & \multicolumn{5}{c|}{Base learning rate   ${\eta}$}             \\ \cline{2-6}
		& 1e-5           & 1e-4  & 1e-3           & 1e-2           & 1e-1  \\ \hline
		Houston & 71.79 & 78.45          & 79.74 & \textbf{80.25} & 66.33 \\ \hline
		Pavia   & 82.41 & 82.25 & 80.5  & \textbf{84.83} & 75.99 \\ \hline
		GID     & 59.11 & 71.23          & 70.3  & \textbf{78.69} & 74.65 \\ \hline \hline
	\end{tabular}
\vspace{-0.1em}
\end{table}

\begin{table}[]
	\caption{\label{tab:lambda}
		Overall classification accuracy (\%) with different regularization parameters ${\lambda}$ for the proposed LDGnet using the three experimental data.}
	\centering
	\begin{tabular}{|c|c|c|c|c|c|}
		\hline\hline
		\multirow{2}{*}{Target scene} & \multicolumn{5}{c|}{Regularization parameter ${\lambda}$}             \\ \cline{2-6}
		& 1e-3           & 1e-2  & 1e-1           & 1e+0           & 1e+1  \\ \hline
		Houston & 75.84 & 73.82          & 77.49 & \textbf{80.25} & 78.44 \\ \hline
		Pavia   & 81.58 & \textbf{84.83} & 82.09 & 79.66          & 81.87 \\ \hline
		GID     & 73.21 & 70.44          & 75.15 & \textbf{78.69} & 73.02 \\ \hline \hline
	\end{tabular}
	\vspace{-0.1em}
\end{table}

\begin{table}[]
	\caption{\label{tab:alpha}
		Overall classification accuracy (\%) with different weights ${\alpha}$ for the proposed LDGnet using the three experimental data.}
	\centering
	\begin{tabular}{|c|c|c|c|c|c|}
		\hline\hline
		\multirow{2}{*}{Target scene} & \multicolumn{5}{c|}{ Weight ${\alpha}$}             \\ \cline{2-6}
		& 	0.1            & 0.3            & 0.5   & 0.7  & 0.9   \\ \hline
		Houston & 79.81          & \textbf{80.25} & 78.98 & 78.6 & 77.16 \\ \hline
		Pavia   & 80.29          & \textbf{84.83} & 82.53 & 81.9 & 80.93 \\ \hline
		GID     & \textbf{78.69} & 72.91          & 77.13 & 70.6 & 77.62 \\ \hline\hline
	\end{tabular}
	\vspace{-0.1em}
\end{table}


\begin{table}[tp]
	\begin{center}
		\centering
		\setlength{\tabcolsep}{1mm}
		\caption{\label{table:Ablation}
			Ablation comparison of each variant of LDGnet.}
		\begin{tabular}{|c|cccc|}
			\hline
			\hline
		\multicolumn{1}{|c|}{Model}    & \multicolumn{1}{c|}{\begin{tabular}[c]{@{}c@{}}LDGnet (cls)\end{tabular}}            & \multicolumn{1}{c|}{\begin{tabular}[c]{@{}c@{}}LDGnet (coarse)\end{tabular}} & \multicolumn{1}{c|}{\begin{tabular}[c]{@{}c@{}}LDGnet (fine)\end{tabular}} & \multicolumn{1}{c|}{\begin{tabular}[c]{@{}c@{}}LDGnet\end{tabular}} \\ \hline
		\multicolumn{1}{|c|}{Data set} & \multicolumn{4}{c|}{Houston}
		\\ \hline
		OA (\%)                        & \multicolumn{1}{c|}{74.99} & \multicolumn{1}{c|}{78.35}                                  & \multicolumn{1}{c|}{79.80}                                               & \multicolumn{1}{c|}{\textbf{80.25}}                   \\ \hline
		KC ($\kappa$)                  & \multicolumn{1}{c|}{61.28} & \multicolumn{1}{c|}{65.31}                                  & \multicolumn{1}{c|}{66.48}                & \multicolumn{1}{c|}{\textbf{65.68}}                                \\ \hline
		& \multicolumn{4}{c|}{Pavia}                                                                                                                                                                                                                                                                                            \\ \hline
		OA (\%)                        & \multicolumn{1}{c|}{75.41} & \multicolumn{1}{c|}{83.32}                                    & \multicolumn{1}{c|}{81.18}              & \multicolumn{1}{c|}{\textbf{84.83}}       \\ \hline
		KC ($\kappa$)                  & \multicolumn{1}{c|}{70.84} & \multicolumn{1}{c|}{79.96}                                   & \multicolumn{1}{c|}{77.44}               & \multicolumn{1}{c|}{\textbf{81.73}}             \\ \hline
		& \multicolumn{4}{c|}{GID}                                                                                                                                                                                                                                                                                                 \\ \hline
            
		OA (\%)                        & \multicolumn{1}{c|}{72.72} & \multicolumn{1}{c|}{74.18}                                 & \multicolumn{1}{c|}{75.01}               & \multicolumn{1}{c|}{\textbf{78.69}}      \\ \hline
		KC ($\kappa$)                  & \multicolumn{1}{c|}{64.01} & \multicolumn{1}{c|}{66.06}                            & \multicolumn{1}{c|}{67.10}                      & \multicolumn{1}{c|}{\textbf{71.82}}          \\
				\hline
		\hline
	\end{tabular}
\end{center}
\vspace{-0.1em}
\end{table}

\begin{table*}[tp]
	\caption{\label{tab:accuracy_Hou}
		Class-specific and overall classification accuracy (\%) of different methods for the target scene Houston 2018 data.}
	{ 
		\begin{center}
	\begin{tabular}{|c||c|c|c|c|c|c|c|c|}
		\hline
		\hline
		\multirow{2}{*}{Class}  &  \multicolumn{8}{|c|}{Classification algorithms} \\
		\cline{2-9}
		{~}  & {DAAN \cite{yu2019transfer}}&  {MRAN \cite{ZHU2019214}} & {DSAN \cite{2020Deepsub}} & {HTCNN \cite{2019Heterogeneous}}  & {PDEN \cite{li2021progressive}}  & {LDSDG \cite{wang2021learning}}  & {SagNet \cite{nam2021reducing}}  & {LDGnet} \\
				\hline	
			1 & 68.29 & 41.02 & 62.31 & 11.83 & 46.49 & 10.13 & 25.79 & 61.71          \\
			2                       & 77.80  & 76.94 & 77.50  & 70.11 & 77.60  & 62.97 & 62.79 & 77.45          \\
			3                       & 67.50  & 65.91 & 74.55 & 54.99 & 59.73 & 60.81 & 48.66 & 62.08          \\
			4                       & 100   & 100   & 100   & 54.55 & 100   & 81.82 & 81.82 & 95.45            \\
			5                       & 47.69 & 36.90  & 73.39 & 55.60  & 49.62 & 45.65 & 59.57 & 69.53          \\
			6                       & 79.49 & 82.68 & 86.84 & 92.85 & 84.98 & 89.22 & 89.28 & 91.57         \\
			7                       & 45.12 & 56.43 & 46.33 & 46.47 & 64.21 & 44.15 & 34.99 & 45.42          \\	
				\hline
				\multicolumn{1}{|c|}{OA (\%)}& 71.13 & 72.48 & 78.52 & 77.42 & 75.98 & 73.55 & 73.64 & \textbf{80.25}  \\
				\hline
				\multicolumn{1}{|c|}{KC ($\kappa$)}  & 54.93 & 55.83 & 64.45 & 59.94 & 56.12 & 55.17 & 55.32 & \textbf{65.68}\\
				\hline \hline				
			\end{tabular}
	\end{center}}
\vspace{-0.1em}
\end{table*}

\begin{table*}[tp]
	\caption{\label{tab:accuracy_UP}
		Class-specific and overall classification accuracy (\%) of different methods for the target scene Pavia Center data.}
	{ 
		\begin{center}
			\begin{tabular}{|c||c|c|c|c|c|c|c|c|}
				\hline
				\hline
				\multirow{2}{*}{Class}  &  \multicolumn{8}{|c|}{Classification algorithms} \\
				\cline{2-9}
				{~}  & {DAAN \cite{yu2019transfer}}&  {MRAN \cite{ZHU2019214}} & {DSAN \cite{2020Deepsub}} & {HTCNN \cite{2019Heterogeneous}}  & {PDEN \cite{li2021progressive}}  & {LDSDG \cite{wang2021learning}}  & {SagNet \cite{nam2021reducing}}  & {LDGnet} \\
				\hline	
				1 & 71.98 & 59.16 & 93.93 & 96.06 & 85.93 & 91.09 & 98.35 & 95.20          \\
				2                       & 78.98 & 85.15 & 79.80  & 57.70  & 88.56 & 73.51 & 59.76 & 82.79          \\
				3                       & 19.37 & 46.18 & 53.97 & 2.76  & 61.34 & 2.23  & 5.40   & 80.48          \\
				4                       & 58.67 & 69.58 & 75.75 & 93.25 & 85.49 & 71.72 & 87.03 & 85.11          \\
				5                       & 70.87 & 64.58 & 99.44 & 89.94 & 87.95 & 71.04 & 93.19 & 93.15         \\
				6                       & 83.07 & 89.22 & 74.43 & 70.97 & 79.26 & 57.12 & 49.81 & 66.93         \\
				7                       & 55.59 & 60.10  & 67.31 & 42.28 & 64.75 & 78.13 & 57.94 & 81.97          \\	
				\hline
				\multicolumn{1}{|c|}{OA (\%)}& 65.62 & 69.22 & 78.94 & 68.75 & 80.87 & 71.02 & 69.90  & \textbf{84.83}  \\
				\hline
				\multicolumn{1}{|c|}{KC ($\kappa$)}   & 58.85 & 63.35 & 74.90  & 62.60  & 77.02 & 64.62 & 63.44 & \textbf{81.73}\\
				\hline \hline				
			\end{tabular}
	\end{center}}
	\vspace{-0.1em}
\end{table*}

\begin{table*}[tp]
	\caption{\label{tab:accuracy_GID}
		Class-specific and overall classification accuracy (\%) of different methods for the target scene GID-wh data.}
	{ 
		\begin{center}
			\begin{tabular}{|c||c|c|c|c|c|c|c|c|}
				\hline
				\hline
				\multirow{2}{*}{Class}  &  \multicolumn{8}{|c|}{Classification algorithms} \\
				\cline{2-9}
				{~}  & {DAAN \cite{yu2019transfer}}&  {MRAN \cite{ZHU2019214}} & {DSAN \cite{2020Deepsub}} & {HTCNN \cite{2019Heterogeneous}}  & {PDEN \cite{li2021progressive}}  & {LDSDG \cite{wang2021learning}}  & {SagNet \cite{nam2021reducing}}  & {LDGnet} \\
				\hline	
				1 & 93.93 & 36.79 & 94.99 & 27.00    & 79.47 & 22.71 & 37.64 & 72.78         \\
				2                       & 87.67 & 78.93 & 91.33 & 100   & 93.64 & 76.93 & 98.60  & 67.66         \\
				3                       & 11.13 & 74.39 & 11.89 & 0.00     & 2.48  & 99.29 & 1.87  & 64.72         \\
				4                       & 77.85 & 69.06 & 90.21 & 92.63 & 81.91 & 88.13 & 89.40  & 91.59          \\
				5                       & 71.01 & 74.83 & 71.08 & 0.00     & 82.52 & 67.25 & 45.01 & 89.16           \\	
				\hline
				\multicolumn{1}{|c|}{OA (\%)}& 68.06 & 67.49 & 73.69 & 57.20  & 67.71 & 76.48 & 61.64 & \textbf{78.69} \\
				\hline
				\multicolumn{1}{|c|}{KC ($\kappa$)}   & 58.61 & 57.82 & 65.29 & 43.06 & 57.97 & 68.80  & 48.74 & \textbf{71.82}\\
				\hline \hline				
			\end{tabular}
	\end{center}}
	\vspace{-0.1em}
\end{table*}


\begin{figure}[htp]
	\centering
	\setlength{\tabcolsep}{0.5em}
	\begin{tabular}{ccccccccccccc}
		\multicolumn{3}{c}{\begin{sideways}\epsfig{width=0.11\figurewidth,file=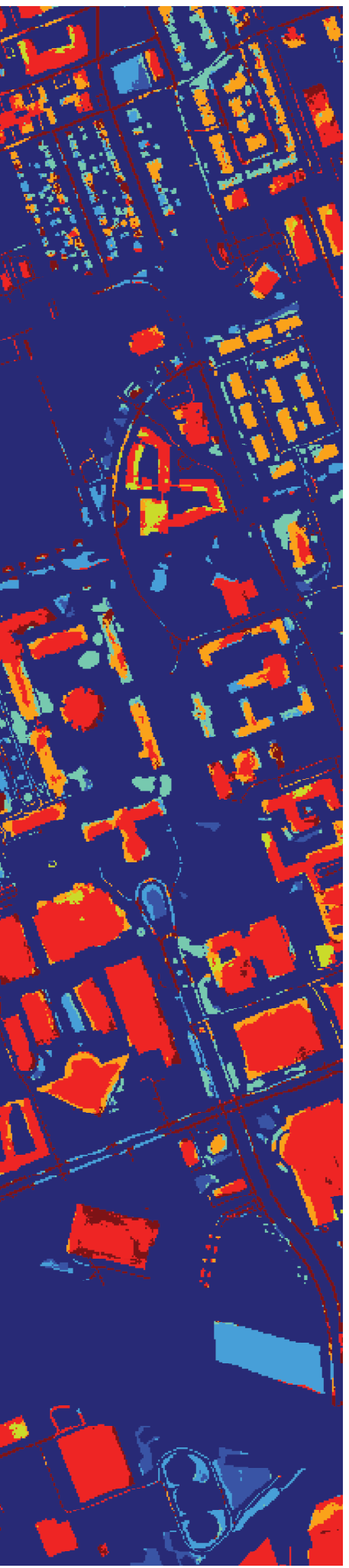}\end{sideways}}     &
		\multicolumn{3}{c}{\begin{sideways}\epsfig{width=0.11\figurewidth,file=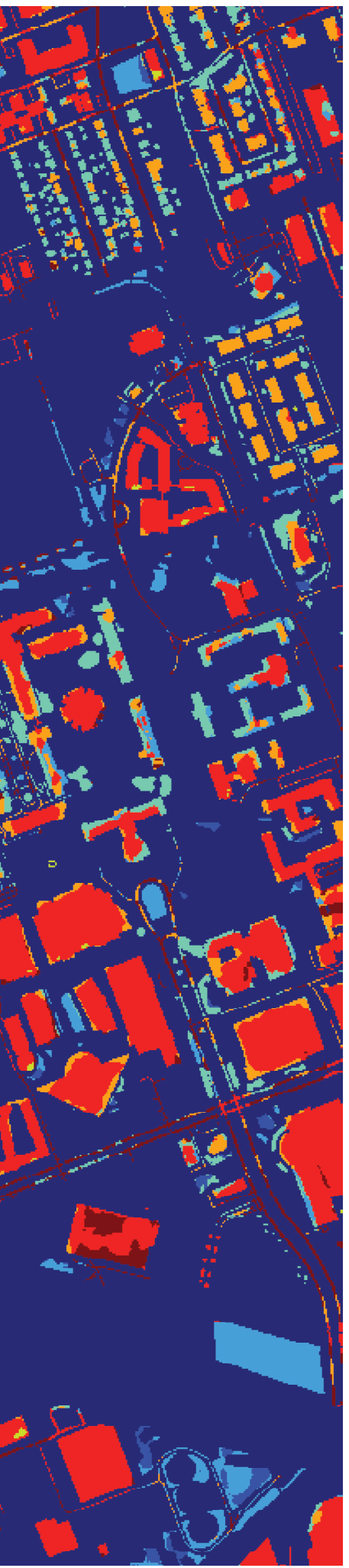}\end{sideways}}  \\
		\multicolumn{3}{c}{(a)} & \multicolumn{3}{c}{(b)} & \\
 		\multicolumn{3}{c}{\begin{sideways}\epsfig{width=0.11\figurewidth,file=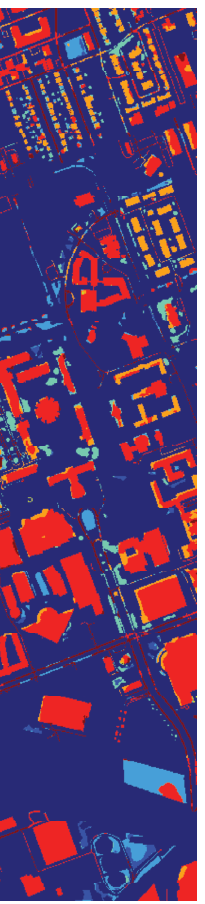}\end{sideways}}    &		\multicolumn{3}{c}{\begin{sideways}\epsfig{width=0.11\figurewidth,file=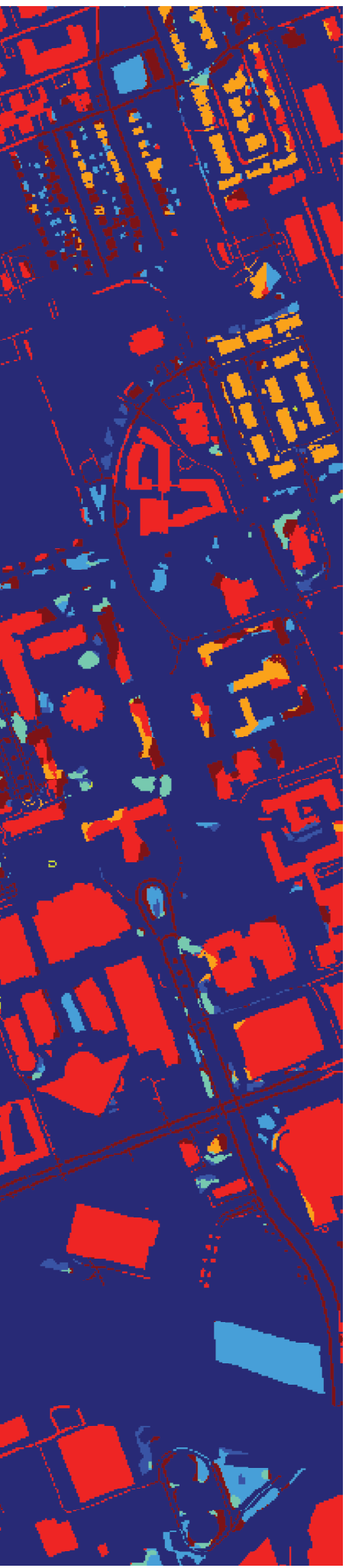}\end{sideways}}     \\
 		\multicolumn{3}{c}{(c)}  & \multicolumn{3}{c}{(d)} &    \\
		\multicolumn{3}{c}{\begin{sideways}\epsfig{width=0.11\figurewidth,file=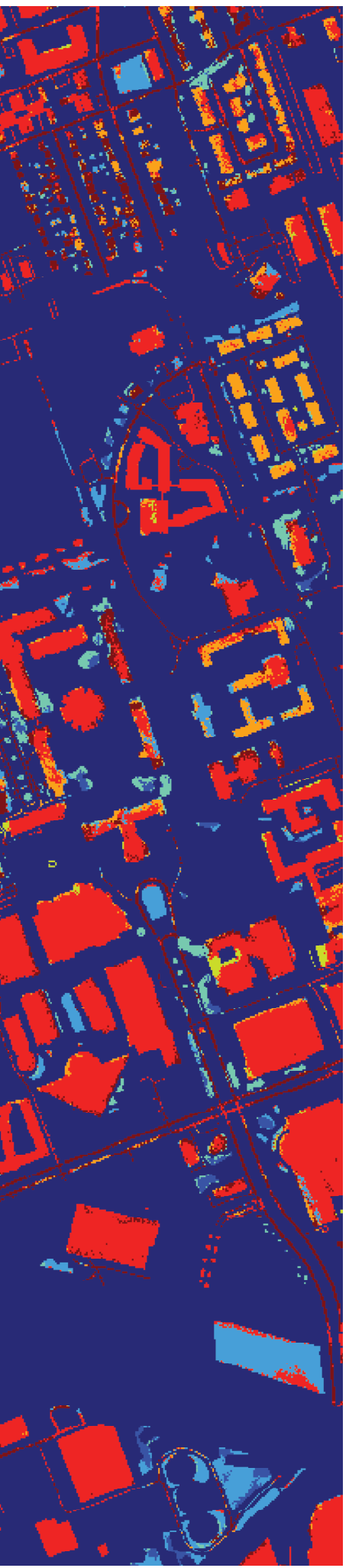}\end{sideways}}     &
		\multicolumn{3}{c}{\begin{sideways}\epsfig{width=0.11\figurewidth,file=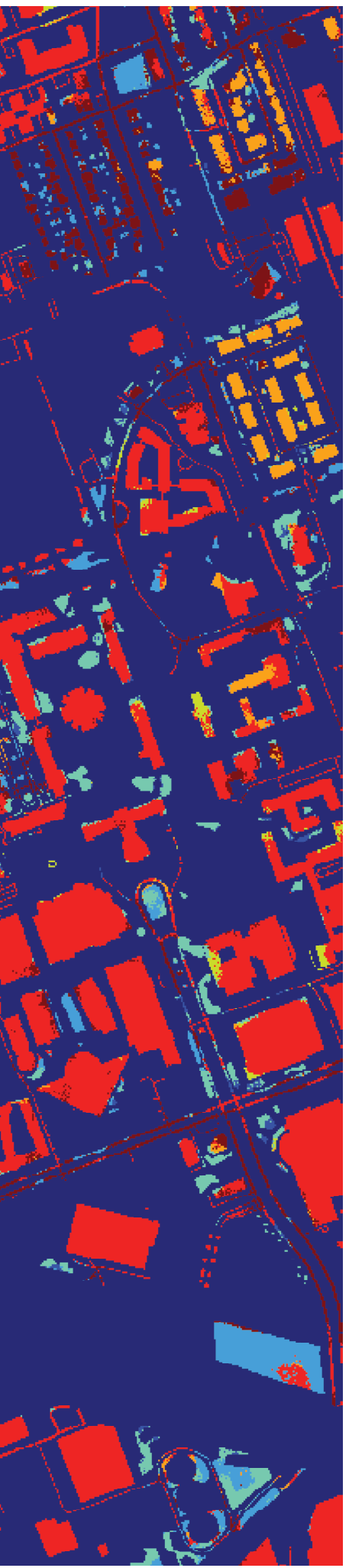}\end{sideways}}  \\
		\multicolumn{3}{c}{(e)} & \multicolumn{3}{c}{(f)} & \\
		\multicolumn{3}{c}{\begin{sideways}\epsfig{width=0.11\figurewidth,file=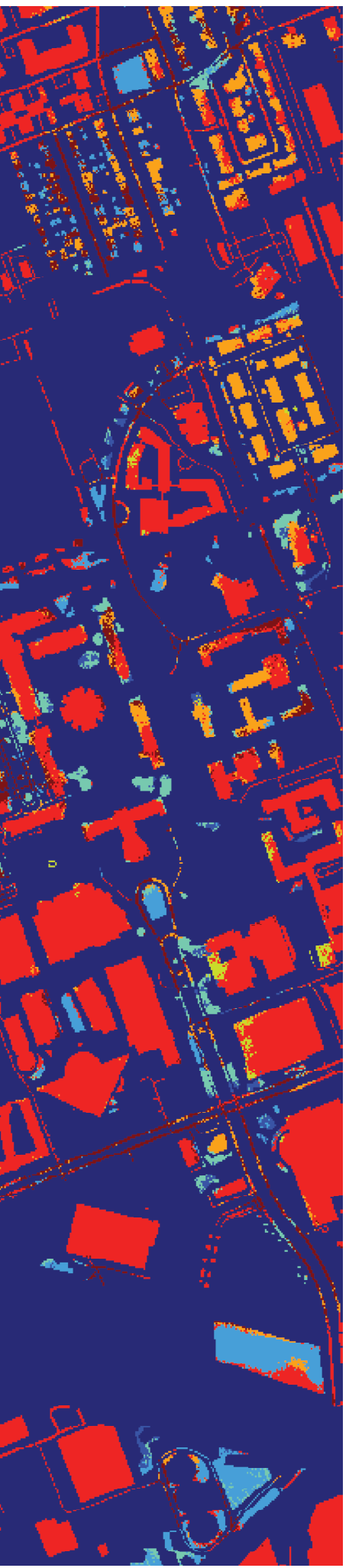}\end{sideways}}    &		\multicolumn{3}{c}{\begin{sideways}\epsfig{width=0.11\figurewidth,file=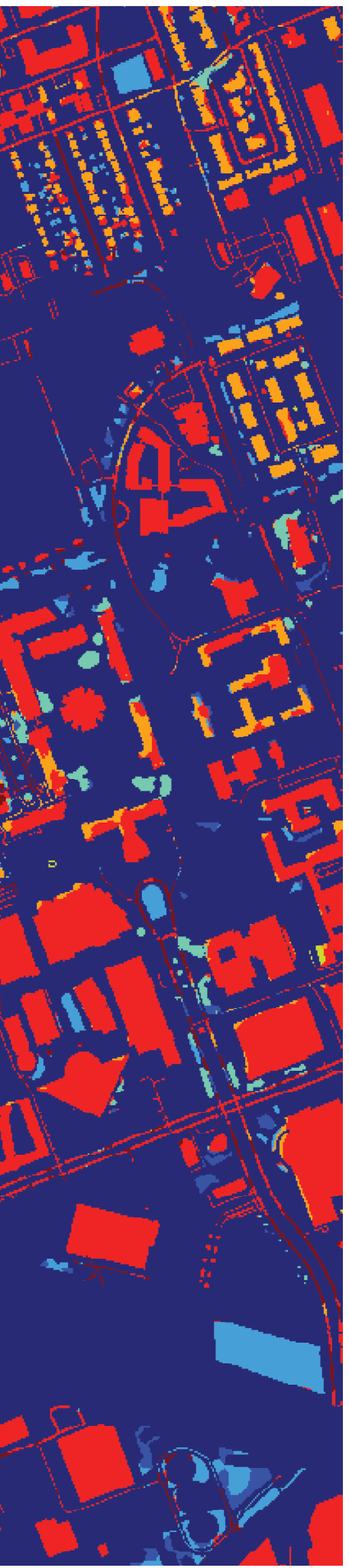}\end{sideways}}     \\
		\multicolumn{3}{c}{(g)}  & \multicolumn{3}{c}{(h)} &    \\
	\end{tabular}
	\vspace{-0.1em}
	\caption{\label{fig:Map_Hou}
		Data visualization and classification maps for target scene Houston 2018 data obtained with different methods including: (a) DAAN (71.13\%), (b) MRAN (72.48\%), (c) DSAN (78.52\%), (d) HTCNN (77.42\%), (e) PDEN (75.98\%), (f) LDSDG (73.55\%), (g) SagNet (73.64\%), (h) LDGnet (80.08\%).  }
\end{figure}

\begin{figure*}[tp]
	\centering
	\begin{tabular}{ccccccccccccccc}
		\multicolumn{2}{c}{\epsfig{width=0.52\figurewidth,file=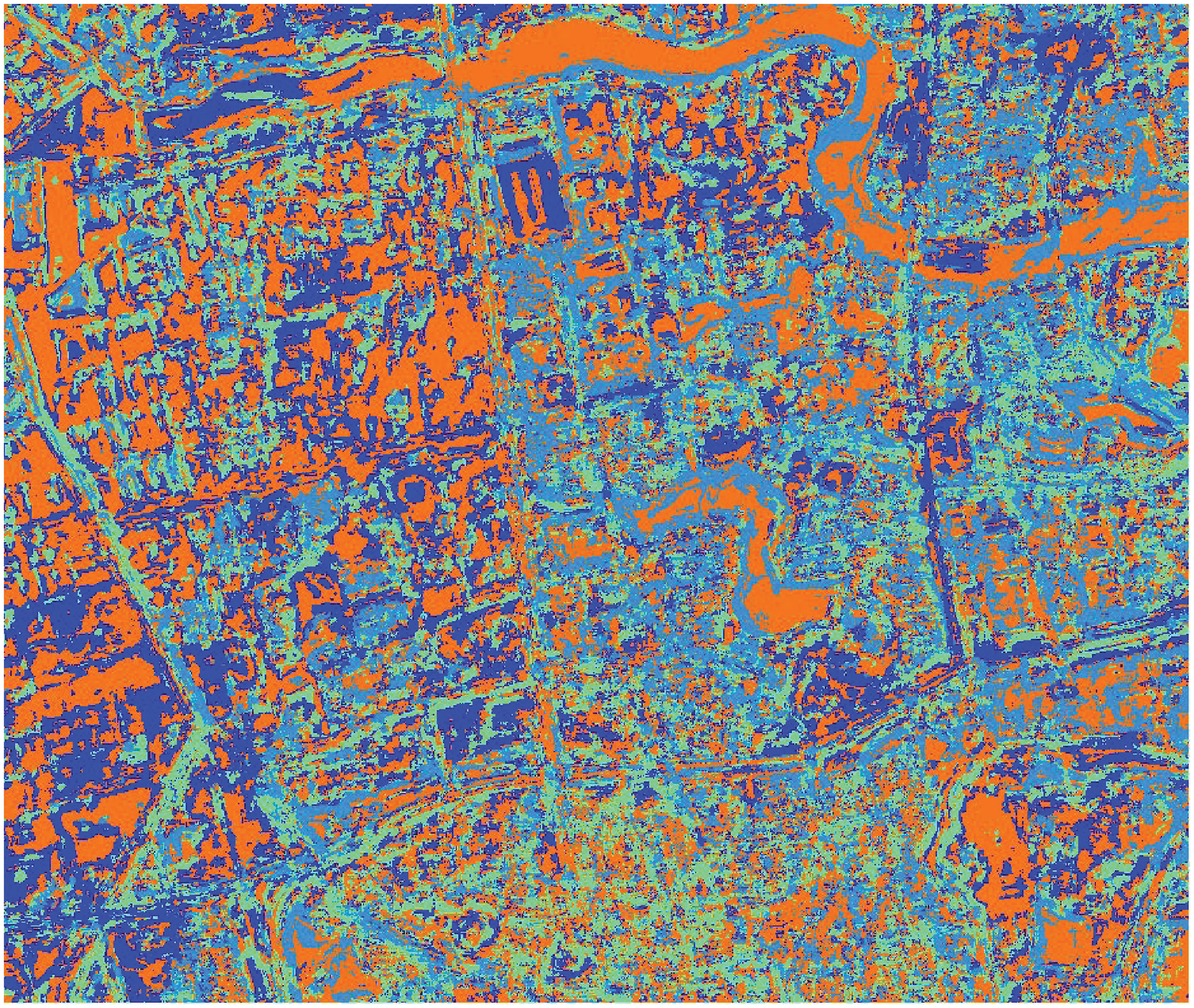}}     &
		\multicolumn{2}{c}{\epsfig{width=0.52\figurewidth,file=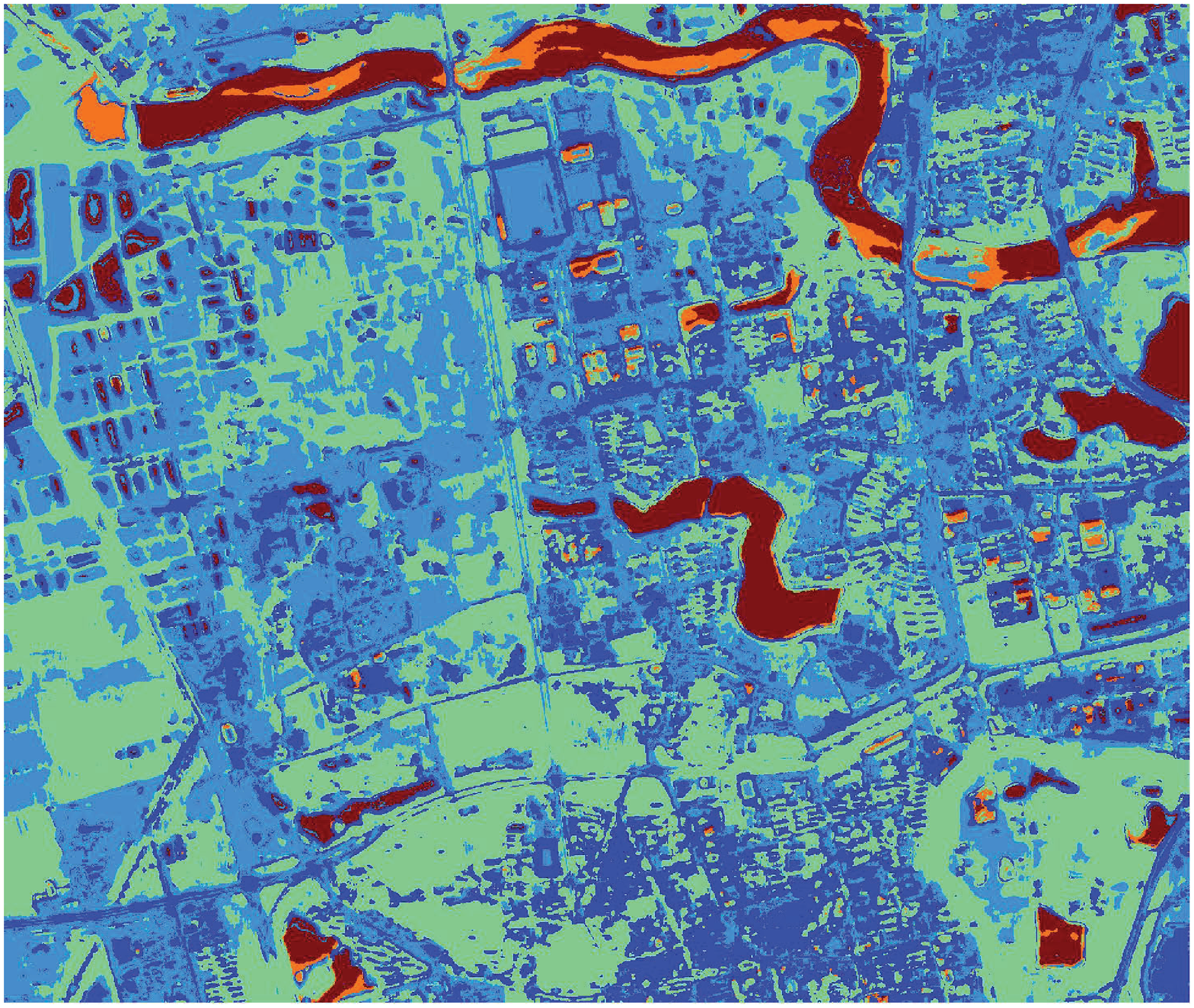}}  & 		\multicolumn{2}{c}{\epsfig{width=0.52\figurewidth,file=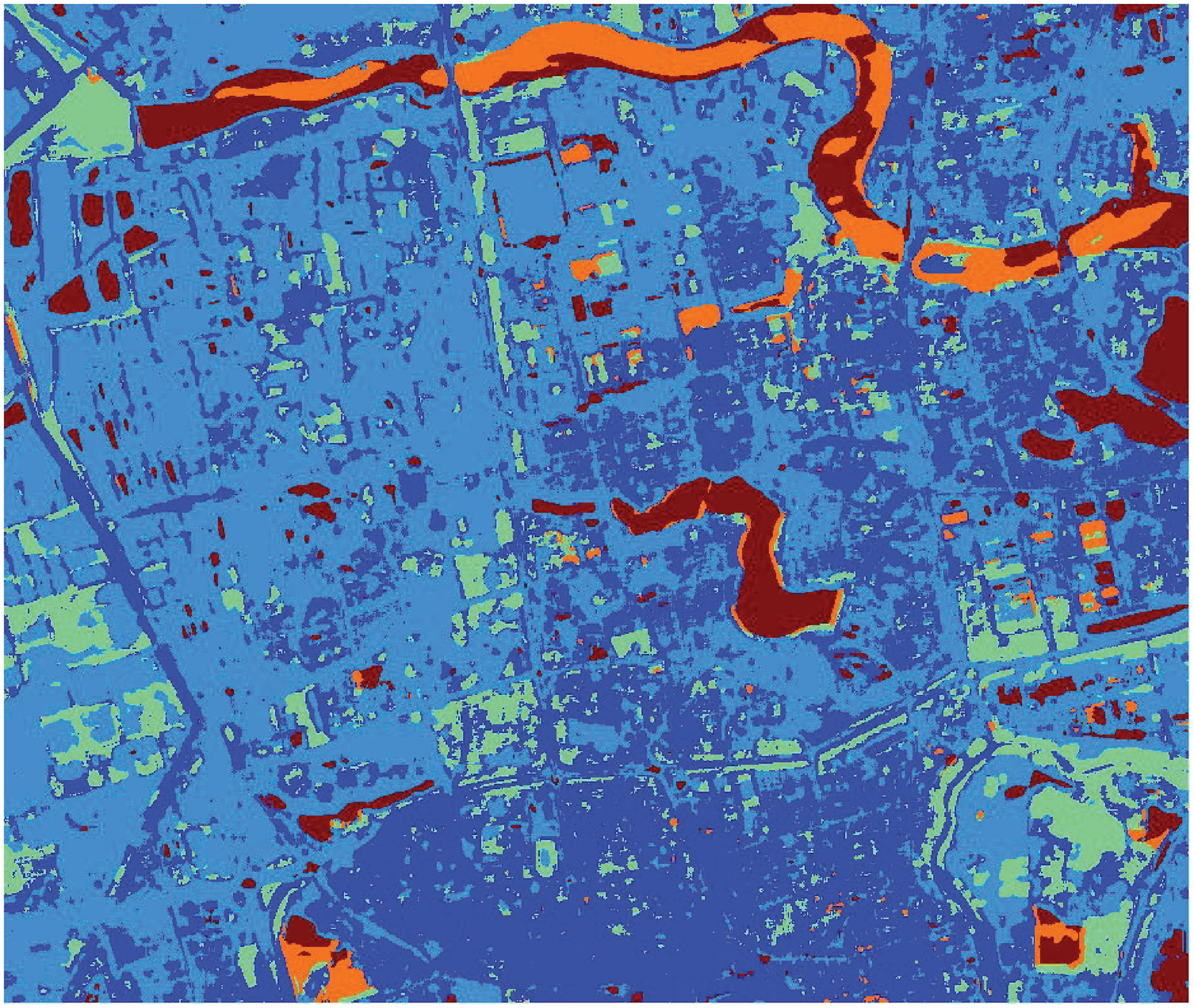}}    &		\multicolumn{2}{c}{\epsfig{width=0.52\figurewidth,file=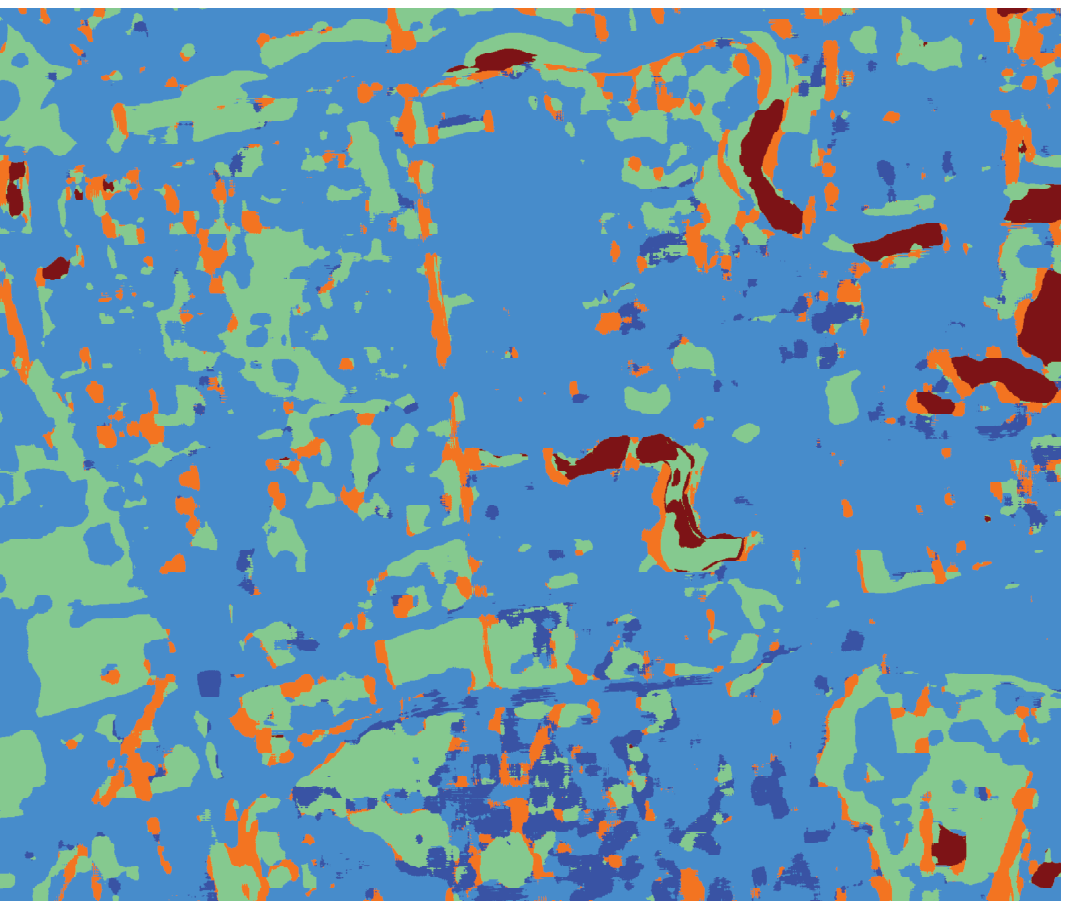}}   \\
		\multicolumn{2}{c}{(a)} & \multicolumn{2}{c}{(b)} & \multicolumn{2}{c}{(c)}  & \multicolumn{2}{c}{(d)}  \\
		\multicolumn{2}{c}{\epsfig{width=0.52\figurewidth,file=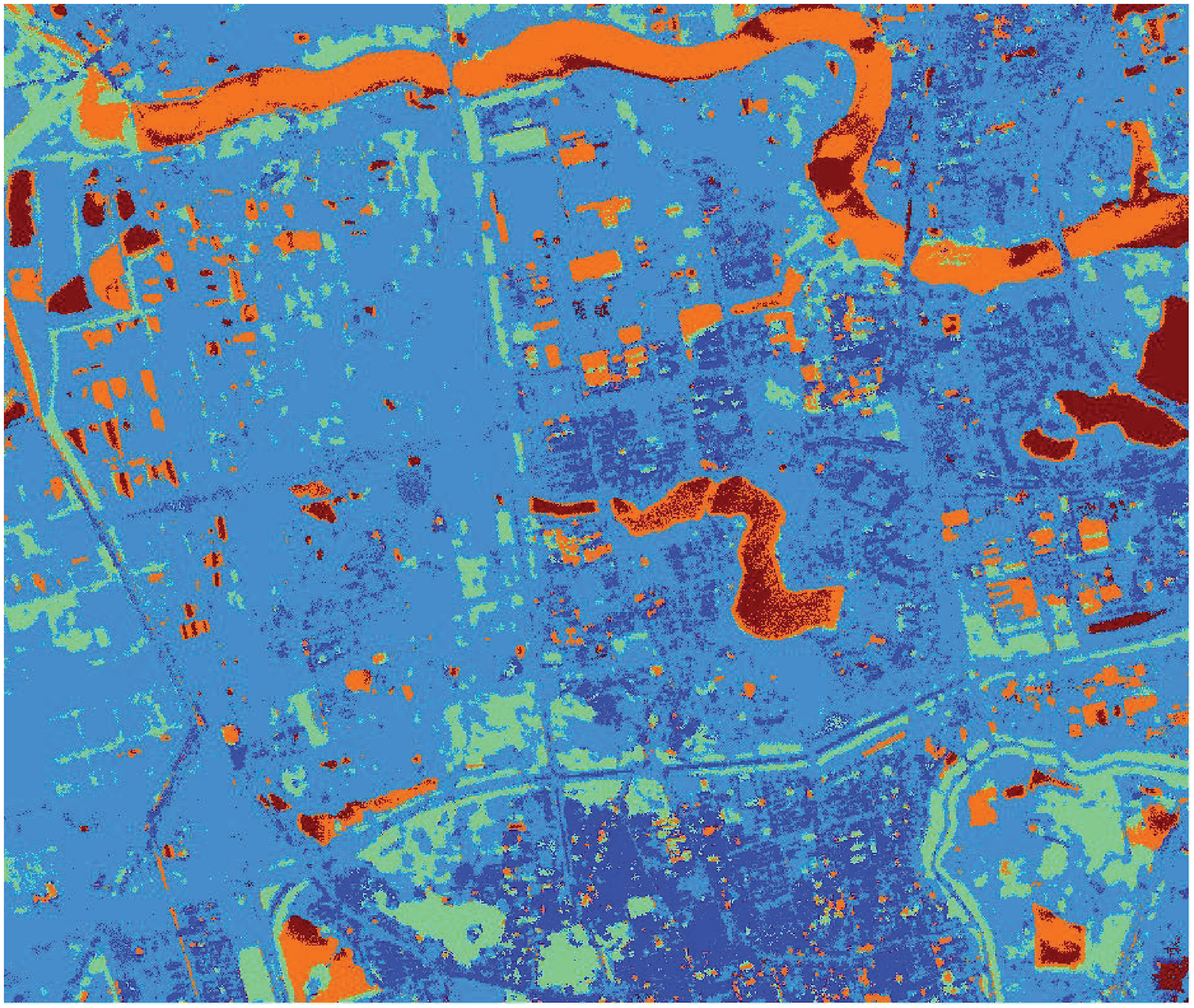}}   &
		\multicolumn{2}{c}{\epsfig{width=0.52\figurewidth,file=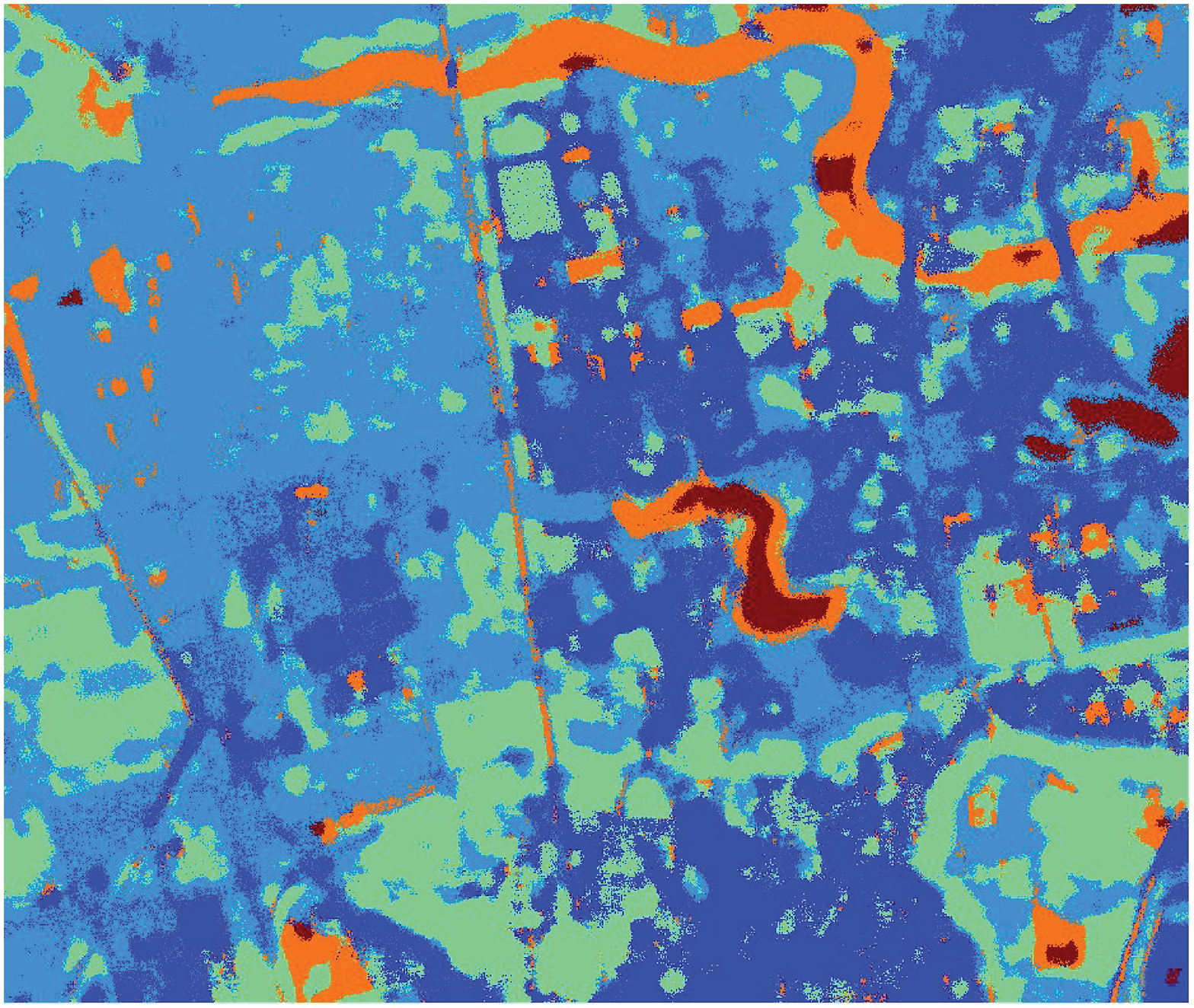}} &
		\multicolumn{2}{c}{\epsfig{width=0.52\figurewidth,file=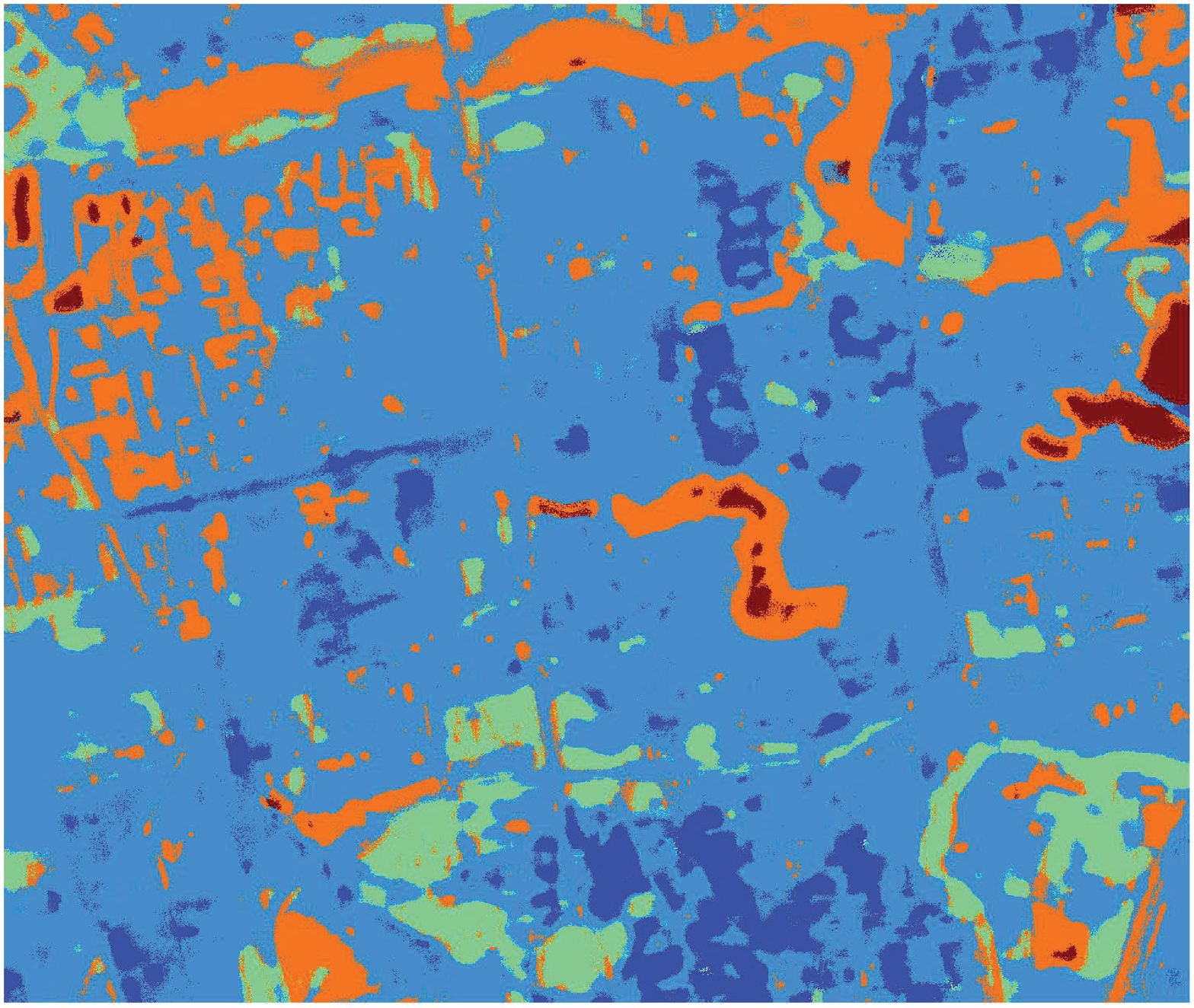}} &
		\multicolumn{2}{c}{\epsfig{width=0.52\figurewidth,file=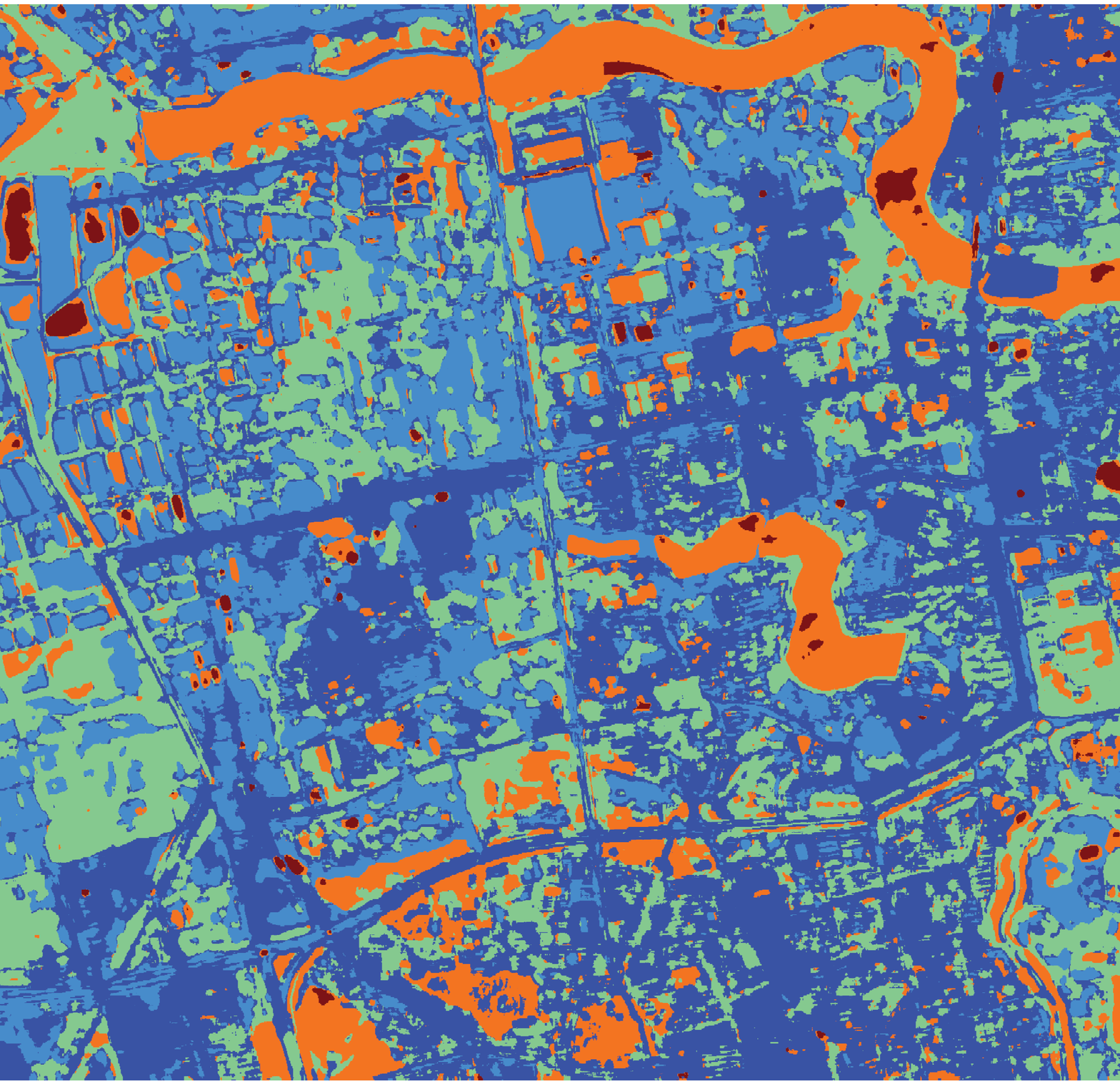}}  \\
		\multicolumn{2}{c}{(e)} &\multicolumn{2}{c}{(f)}& \multicolumn{2}{c}{(g)} & \multicolumn{2}{c}{(h)}\\
	\end{tabular}
	
	\caption{\label{fig:Map_GID}
		Data visualization and classification maps for target scene GID-wh data obtained with different methods including: (a) DAAN (68.06\%), (b) MRAN (67.49\%), (c) DSAN (73.69\%), (d) HTCNN (57.20\%), (e) PDEN (67.71\%), (f) LDSDG (76.48\%), (g) SagNet (61.64\%), (h) LDGnet (77.08\%).  }
\end{figure*}

\begin{table*}[tp]
	\caption{\label{table:text}
		Irrelevant descriptions and relevant descriptions for Trees.}
	\begin{center}
		\begin{tabular}{|c|l|}
			\hline \hline
			Class name                                                                                & \multicolumn{1}{c|}{Fine-grained text}                                                                                    \\ \hline
			& The rigid tell can not crack the garden                                                                                   \\ \cline{2-2} 
			& It was then the formal bus met the imaginary difficulty                                           \\ \cline{2-2} 
			& The heartbreaking profession excites into the abaft guarantee                                                             \\ \cline{2-2} 
			\multirow{-4}{*}{\begin{tabular}[c]{@{}c@{}}Trees\\ (Irrelevant description)\end{tabular}} & The cultural commission sneezes into the federal solution                                                                 \\ \hline
			& In botany, a tree is a perennial plant with an elongated stem, or trunk, usually supporting branches and leaves           \\ \cline{2-2} 
			& Tree is a woody plant that regularly renews its growth                                             \\ \cline{2-2} 
			& Tree has many secondary branches supported clear of the ground on a single main stem or trunk with clear apical dominance \\ \cline{2-2} 
			\multirow{-4}{*}{\begin{tabular}[c]{@{}c@{}}Trees\\ (Relevant description)\end{tabular}}  & Trees appear mostly at the surface and often in clusters with similar height                                              \\ \hline \hline
		\end{tabular}
	\end{center}
\end{table*}

\begin{table*}[tp]
	\caption{\label{table:text_OA}
		Overall classification accuracy (\%) for Irrelevant descriptions and internet descriptions using the Houston 2018 data.}
	\begin{center}
		\begin{tabular}{|c|c|c|c|c|c|c|c|c|c|c|c|}
			\hline \hline
			Irrelevant description & 1 time & 2 time & 3 time & 4 time & 5 time & 6 time & 7 time & 8 time & 9 time & 10 time & mean  \\ \hline 
			OA (\%)                             & 77.96  & 77.20   & 76.91  & 78.08  & 77.04  & 76.82  & 76.02  & 77.86  & 75.99  & 76.46   & 77.03 \\ \hline
			KC ($\kappa$)                            & 59.15  & 57.85  & 57.22  & 59.64  & 57.50   & 56.73  & 55.91  & 58.87  & 55.7   & 56.12   & 57.47 \\ \hline
			Relevant description  & 1 time & 2 time & 3 time & 4 time & 5 time & 6 time & 7 time & 8 time & 9 time & 10 time & mean  \\ \hline
			OA (\%)                             & 79.67  & 78.37  & 78.21  & 78.47  & 78.74  & 78.18  & 81.44  & 79.55  & 80.10   & 79.73   & 79.25 \\ \hline
			KC ($\kappa$)                            & 64.53  & 60.65  & 60.08  & 61.31  & 63.87  & 59.94  & 68.84  & 64.11  & 65.32  & 64.84   & 63.35 \\ \hline \hline
		\end{tabular}
	\end{center}
\end{table*}

\begin{table*}[tp]
	\vspace{-2em}
	\caption{\label{table:time}
		The execution time (in seconds) of one epoch training in different methods.}
	\begin{center}
		\begin{tabular}{|c|c|c|c|c|c|c|c|c|}
			\hline\hline
			Methods    & {DAAN \cite{yu2019transfer}}&  {MRAN \cite{ZHU2019214}} & {DSAN \cite{2020Deepsub}} & {HTCNN \cite{2019Heterogeneous}}  & {PDEN \cite{li2021progressive}}  & {LDSDG \cite{wang2021learning}}  & {SagNet \cite{nam2021reducing}}  & {LDGnet} \\ \hline
			Houston 2018 & 15.68 & 13.52 & 16.64 & 31.21 & 4.88  & 43.56  & 7.03   & 17.84   \\ \hline
			Pavia Center & 28.32 & 29.01 & 31.76 & 46.24 & 14.21 & 63.38 & 20.44  & 68.21  \\ \hline
			GID-wh       & 20.35 & 22.32 & 24.43 & 39.65 & 3.15  & 20.78  & 16.78  & 19.53   \\ \hline \hline
		\end{tabular}
	\end{center}
\vspace{-2em}
\end{table*}

\subsection{Parameter Tuning}

A parameter sensitivity analysis is conducted to evaluate the sensitivity of LDGnet on the three TDs. The base learning rate ${\eta}$, regularization parameter $\lambda$ and weight $\alpha$, regarded as adjustable hyperparameters are selected from \{$1e-5$, $1e-4$, $1e-3$, $1e-2$, $1e-1$\}, \{$1e-3$, $1e-2$, $1e-1$, $1e+0$, $1e+1$\} and \{0.1, 0.3, 0.5, 0.7, 0.9\}, respectively. 

After being modified by the learning rate, the gradient of loss function is employed in the gradient descent updates to estimate the model weight hyperparameters. Table \ref {tab:lr} provides classification results corresponding to different base learning rates in three data sets. With respect to three datasets, $1e-2$ is the ideal base learning rate. Table \ref {tab:lambda} and Table \ref {tab:alpha} include the OA of all experimental datasets for regularization parameter $\lambda$ and weight $\alpha$, respectively. The optimal $\lambda$ is 1e+0 for the Houston dataset and GID dataset, 1e-2 for the Pavia dataset, and $\alpha$ is 0.3 for the Houston dataset and Pavia dataset, and 0.1 for the GID dataset.

\subsection{Ablation Study}

The image encoder and text encoder are the key components of LDGnet, and visual-linguistic alignment is the main strategy for learning domain invariant representation. Ablation analyses are carried out by eliminating each component from the total framework in order to evaluate the contribution of important LDGnet components.

There are three variants in the ablation analyses, (1) ``LDGnet (cls)": the text encoder and projection head are deleted from LDGnet, (2) ``LDGnet (coarse)": the fine-grained text is deleted, and (3) ``LDGnet (fine)": the coarse-grained text is deleted. It is clear from Table \ref{table:Ablation} that the proposed LDGnet performs better than existing variations and makes significant advancements. On the basis of LDGnet (cls), add transformer as the text encoder architecture, paired with coarse-grained text descriptions LDGnet (coarse) or fine-grained text descriptions LDGnet (fine), and performance improves by around 6\%. This suggests that linguistic modality aids in the learning of visual representations, and both coarse- and fine-grained visual-linguistic alignment can help to improve the generalization ability of the model on TD. Furthermore, LDGnet performs 1\%$\sim$4\% better than LDGnet (coarse) and LDGnet (fine). The use of coarse-grained or fine-grained linguistic features alone is inferior to the fusion of them, so as to enrich language supervised signals and complete visual-linguistic alignment at two levels.

\subsection{Fine-grained text analysis}

The fine-grained text descriptions provide prior knowledge in the proposed LDGnet. Tables \ref{table:Houston_text}-\ref{table:GID_text} shows the fine-grained text artificially defined from the color, shape, distribution and adjacency relationship. The following experiments are carried out with the Houston dataset as an example to analyze the sensitivity of LDGnet to how fine-grained text is defined: (1) Irrelevant description: four illogical sentences are randomly generated for each land cover class, and these sentences have nothing to do with the attributes of land cover class; (2) Relevant description: the name of each land cover class is taken as the entry, and four sentences corresponding to the description are selected from Google. Table \ref{table:text} shows the irrelevant and Internet descriptions for the Trees used in the experiment. These sentences which are closely related to the land cover class. In the experiment, two of the four fine-grained text descriptions of each class are randomly selected (two identical descriptions may be selected for a class).

Both experiments are run ten times, and the classification accuracy is shown in Table \ref{table:text}. When irrelevant descriptions are used as fine-grained text, it is decreased by 3\% lower compared to 80.25\% using artificially defined descriptions, and by 1\% when relevant Internet descriptions are used. It is clear that relevant fine-grained text descriptions, whether derived from human definitions or Internet searches, perform better than unrelated ones, indicating  that a correct description of the land cover class in the linguistic modality is necessary. Additionally, LDGnet has considerable performance when using relevant fine-grained text from Internet, which indicates that the proposed method is compatible with multiple descriptions of a land cover class, not one defined in Table \ref{table:Houston_text}-\ref{table:GID_text}.

\subsection{Performance on Cross-Scene HSI Classification}
\label{sec:Classification Performance}

Relevant algorithms such as DAAN, MRAN, DSAN, HTCNN, PDEN, LDSDG, and SagNet are utilized for comparison in order to assess the performance of LDGnet with only SD employed for training. The training samples is set as follows. All SD data with labels (80\% for training and 20\% for validation) and all TD data without labels are used for training with DAAN, MRAN, DSAN, and HTCNN, which are considered DA techniques. Only SD with labels (80\% for training and 20\% for validation) are used as training examples for DG techniques, PDEN, LDSDG, and SagNet, and the patch size of LDSDG and SagNet is adjusted to 32$\times$32 to accommodate the input size of Resnet18. In contrast to the other two datasets, the SD in Houston dataset is also increased by four times through random flip and random radiation noise (illumination). The optimal base learning rate and regularization parameters of all comparison algorithms are selected from \{$1e-5$, $1e-4$, $1e-3$, $1e-2$, $1e-1$\} and \{$1e-3$, $1e-2$, $1e-1$, $1e+0$, $1e+1$, $1e+2$\}, respectively, and cross-validation is used to find the corresponding optimal parameters. 
 
The following analyses are obtained from Tables \ref{tab:accuracy_Hou}-\ref{tab:accuracy_GID}.

\begin{itemize}
	\item On all TDs, DSAN exhibits the best performance for the DA technique. In the comparison of DG approaches, PDEN performs well on the data from Houston 2018 and the Pavia Center, while LDSDG works well on the data from GID-wh. In particular, DSAN provides 2\% improvement in OA over PDEN on Houston 2018 data, while PDEN and LDSDG are 2\% higher than DSAN on Pavia Center and GID-wh, respectively. This demonstrates that TD is not always employed in the training process to produce the greatest classification performance and that both the DA method and the DG approach have their own advantages in various scenes.
	
	\item LDGnet outperforms DSAN on all TDs by 4\% to 6\%. DSAN expressly employs the domain alignment technique during the training process to minimize the domain shift and directly accesses TD. In contrast, self-supervised contrastive learning is employed in LDGnet to reduce the gap between visual and linguistic features by class while learning domain invariant representation in cross-domain shared semantic space. The enhanced LDGnet classification performance demonstrates the superiority of this explicit shared space learning technique over the domain alignment strategy used in DA.
	
	\item LDGnet outperforms the DG algorithms, PDEN and LDSDG et al., by 2\% to 6\% on OA.  Different from data generation strategies in PDEN and LDSDG, coarse-grained and fine-grained text representations are introduced in LDGnet to learn more robust domain invariant representations from multi-modal features through visual-linguistic alignment.
\end{itemize}

Classification maps are illustrated in Figs. \ref{fig:Map_Hou}-\ref{fig:Map_GID}. In Figs. \ref{fig:Map_Hou}, labeled pixels are displayed as ground truth and unlabeled pixels as backgrounds, and all pixels are predicted for comparison in Fig. \ref{fig:Map_GID}. In contrast, the proposed LDGnet obtains less noisy and more accurate results in some areas of the classification maps, such as 3-rd class (Brick) and 7-th bare soil in Pavia Center data are greatly improved compared to all comparison methods. It is obvious from Fig. \ref{fig:Map_GID} that the 3-rd (Garden land) and 5-th (Lake) in GID-wh data are better predicted.

The one epoch training time on all experimental data are presented in Table \ref{table:time} in order to illustrate the computational difficulty of various approaches. All the experiments are carried out using Pytorch on an AMD EPYC 7542 32-Core Processor (48-GB RAM) powered with Nvidia GTX 3090 GPU with 24GB memory. The computational cost of LDGnet is not low compared with other methods, and the three-layer transformer used as the text encoder is the primary source of calculation expense. Its training parameters 33.43M accounted for 97.7\% of the total number of parameters. The overall computational complexity of LDGnet is lower than that of LDSDG, where the Style-Complement modul is designed to consider multiple potential style changes.

%% file: conclusions.tex
The Language-aware Domain Generalization Network (LDGnet), which combines the visual and language modes, is proposed. The visual and linguistic features are extracted concurrently using the dual-stream architecture. The text representation with prior knowledge is regarded as cross-domain shared knowledge, which is designed to guide the learning of cross-domain invariant representation. Specifically, the coarse-grained and fine-grained text representations are designed, using CNN-based image encoder and transformer based text encoder to extract visual features, coarse-grained and fine-grained linguistic features, respectively. The semantic space constituted of linguistic features is treated as a cross-domain shared space, and supervised contrastive learning is used to gradually decrease the gap between visual and linguistic features in order to achieve visual-linguistic alignment. Finally, for target domain scene generalization, the image encoder embedded with common language knowledge is applied. Comprehensive experiments on three datasets validate the effectiveness of the proposed LDGnet in domain generalization.